\documentclass[a4paper,fleqn]{cas-dc}

\usepackage[numbers]{natbib}
\usepackage{scrextend}
\usepackage{graphicx}
\usepackage{xcolor}
\usepackage{amssymb}
\usepackage{multirow}
\usepackage{rotating}
\usepackage{url}
\usepackage[utf8]{inputenc}
\usepackage[T1]{fontenc}
\usepackage[final]{microtype}
\usepackage{balance}
\inputencoding{utf8}

\def\tsc#1{\csdef{#1}{\textsc{\lowercase{#1}}\xspace}}
\tsc{WGM}
\tsc{QE}
\tsc{EP}
\tsc{PMS}
\tsc{BEC}
\tsc{DE}

\begin{document}
\let\WriteBookmarks\relax
\def\floatpagepagefraction{1}
\def\textpagefraction{.001}
\shorttitle{Conceptual Framework and Review of XRL}
\shortauthors{R Dazeley et~al.}



\title [mode = title]{Explainable Reinforcement Learning for Broad-XAI: A Conceptual Framework and Survey} 

\newcommand{\Authornote}[2]{\small\textcolor{red}{\sf$[${#1: #2}$]$}}
\newcommand{\sanote}[2]{{\Authornote{Sunil}{#1}}} 
\newcommand{\rpdnote}[2]{{\Authornote{Richard}{#1}}} 



\author[1]{Richard Dazeley}[
                        orcid=0000-0002-6199-9685]
\cormark[1]
\ead{richard.dazeley@deakin.edu.au}

\credit{Conceptualization of this study, Methodology, Software}

\address[1]{School of Information Technology, Deakin University, Locked Bag 20000, Geelong, Victoria 3220, Australia}

\author[2]{Peter Vamplew}[orcid=0000-0002-8687-4424]
\ead{p.vamplew@federation.edu.au}
\author[1]{Francisco Cruz}[orcid=0000-0002-1131-3382]
\ead{francisco.cruz@deakin.edu.au}

\address[2]{School of Science, Engineering and Information Technology, Federation University Australia, Ballarat, Victoria 3353, Australia}

\cortext[cor1]{Corresponding author}


\begin{abstract}
Broad Explainable Artificial Intelligence (\textit{Broad-XAI}) moves away from interpreting individual decisions based on a single datum and aims to provide integrated explanations from multiple machine learning algorithms into a coherent explanation of an agent’s behaviour that is aligned to the communication needs of the explainee. Reinforcement Learning (RL) methods, we propose, provide a potential backbone for the cognitive model required for the development of Broad-XAI. RL represents a suite of approaches that have had increasing success in solving a range of sequential decision-making problems. However, these algorithms all operate as black-box problem solvers, where they obfuscate their decision-making policy through a complex array of values and functions. EXplainable RL (XRL) is relatively recent field of research that aims to develop techniques to extract concepts from the agent's: perception of the environment; intrinsic/extrinsic motivations/beliefs; Q-values, goals and objectives. This paper aims to introduce a conceptual framework, called the Causal XRL Framework (CXF), that unifies the current XRL research and uses RL as a backbone to the development of Broad-XAI. Additionally, we recognise that RL methods have the ability to incorporate a range of technologies to allow agents to adapt to their environment. CXF is designed for the incorporation of many standard RL extensions and integrated with external ontologies and communication facilities so that the agent can answer questions that explain outcomes and justify its decisions. This paper aims to: establish XRL as a distinct branch of Explainable Artificial Intelligence (XAI); introduce a conceptual framework for XRL;  review existing approaches explaining agent behaviour; and identify opportunities for future research. Finally, this paper will discuss how additional information can be extracted and ultimately integrated into models of communication, facilitating the development of Broad-XAI. 

\end{abstract}




\maketitle

\section{Introduction}
\label{Sec: Introduction}

Successes, such as AlphaGo \cite{silver2016mastering}, autonomous vehicles \cite{huval2015empirical} and playing atari video games \cite{mnih2015human}, saw the MIT Technology Review list Reinforcement Learning (RL) as one of the top ten technologies of 2017 \cite{knight2017breakthrough}. However, while RL can often solve complex sequential decision-making problems, the algorithms currently operate as a black-box, where experts must analyse vast amounts of data and functions to determine why they make particular decisions. For example, during AlphaGo's second challenge against Lee Sedol (ranked 9-dan) AlphaGo's 37th turn surprised both commentators and Lee Sedol, which turned the course of the game in AlphaGo's favour \cite{Metz2016intwomoves}. David Silver, DeepMind researcher, reportedly had no insight into why AlphaGo made such a creative move until he had investigated the actual calculations made by the program \cite{Metz2016deepmind}. For these systems to go the next step and be used by everyday non-expert users that are not able to inspect an agent's internal representation of its policy, they must be able to provide explanations for their behaviour.

Investigations into the development of explanation facilities for Artificial Intelligence (AI) have progressed steadily for nearly fifty years. For instance, Knowledge-Based Systems (KBS) researchers and designers such as \citeauthor{shortliffe1975model} \cite{shortliffe1975model} first discussed approaches to providing explanations in the mid-70s through rule tracing. These were later applied in a range of early projects such as \citeauthor{davis1977production} \cite{davis1977production}, \citeauthor{swartout1983xplain} \cite{swartout1983xplain} and \citeauthor{chandrasekaran1988explanation} \cite{chandrasekaran1988explanation}. These early ideas were later extended in domains such as Bayesian networks \cite{lacave2002review}, early neural network systems \cite{andrews1995survey} and recommender systems \cite{cramer2008effects, assad2007personisad}. 

The last decade has seen a surge in the uptake of machine learning systems, while at the same time the systems have become increasingly obfuscated and non-transparent to end-users. This has resulted in significant growth in research, and funding, for the development of eXplainable Artificial Intelligence (XAI) and Interpretable Machine Learning (IML)\footnote{The literature also uses the term Explainable Machine Learning (XML) interchangeably with IML. As most of this work is also focused on the interpretation of algorithms' decision-making this paper will refer to all this work as IML.} \cite{abdul2018trends}. Drivers for this increase include: increased social anxiety towards automated systems; increases in funding such as Defense Advanced Research Projects Agency (DARPA) \cite{gunning2018DARPA}; legislative changes such as the European Union's new General Data Protection Regulation \cite{goodman2016european}; and, increased interest from futurists and start-up companies like AGI Innovations \cite{voss2018AGIInnovations} and bons.ai \cite{hammond2018bonsai}.

The majority of current research has focused on the interpretation of an AI's decision. \citeauthor{miller2017explainable} \cite{miller2017explainable}, emphasizes this view, stating that AI researchers typically build explainability from their perspective resulting in mostly interpretable or debugging like explanations. However, for AI to succeed, it must provide trusted and socially acceptable systems and, therefore, should be modelled on philosophical, psychological and cognitive science models of human explanation. \citeauthor{dazeley2021levels} \cite{dazeley2021levels} identified a set of levels of explainability that are adapted from Animal Cognitive Ethology’s levels of intentionality \cite{griffin1976animalawareness, chen2017survey}, combined with human social contexts \cite{dazeley2008epistemological}. After reviewing the literature across these levels, \citeauthor{dazeley2021levels} \cite{dazeley2021levels} shows that the majority of XAI research is focused on the lowest level of Zero-order explanations. Furthermore, that a fully \textit{Broad-XAI} system requires the full range of XAI levels to provide an integrated conversational explanation. While \citeauthor{dazeley2021levels} \cite{dazeley2021levels} argues that a mix of technologies can be used to develop broad-XAI, RL approaches were identified as one approach for providing a backbone supporting explanations across multiple levels of explainability.

Research into explaining a decision made by a RL-based agent has been increasing with several papers being published recently. Occasionally this work is subsumed under the banner of XAI and IML. XAI covers explanation of any AI based decision and, therefore, encompasses a broad range of decision-making situations --- including those conducted in an RL agent. 
However, explainability for an RL agent, while clearly a subset of XAI and with similarities to IML, has distinct characteristics that requires its explicit separation from current XAI and IML research. The term eXplainable Reinforcement Learning (XRL) has begun to emerge recently to cover research into explaining agent's decisions during temporally separated decision-making tasks. Separating explanation for RL from general XAI and IML provides the opportunity for RL researchers to more easily identify potential avenues of research and to track developments for providing explanations from an RL agent. 

Considering explanation in an RL context allows extra focus on the types of explanations possible in RL and how they can be combined to provide levels of explanation to better facilitate understanding and acceptance by different users. \citeauthor{heuillet2020explainability} \cite{heuillet2020explainability} and \citeauthor{wallkotter2021explainable} \cite{wallkotter2021explainable} have both provided in-depth surveys of the issues and abilities that reinforcement learning and embodied agents can provide. These papers pull together a number of papers that have explored the potential of explainable systems in interactive temporal agents. In this paper, we aim to go beyond exploring the current work alone and instead put forward a conceptual framework that sets up a structure for providing Broad-XAI. The objective is to promote the research and development of systems that can explain behaviour from integrated systems built on a foundation of RL. 
Interactive temporal agents built on this framework would be able to explain decisions and outcomes that provide for the three key areas of human explanation identified by \citeauthor{miller2017explanationBook} \cite{miller2017explanationBook}: contrastive explanation, attribution theory and explanation selection. This framework will be tied to an accepted psychological model of explanation that allows for user controlled and conversational levels of explanation, as discussed by \citeauthor{dazeley2021levels} \cite{dazeley2021levels}. In so doing this paper suggests that, if RL decisions can be explained using human models of explanation, then they can build more trust and social acceptance. In presenting this framework, this paper will discuss plausible approaches to developing each component, as well as identify current work in each area. 

This paper is structured with six further parts. The next section will provide a background to XAI and argue how XRL presents a distinct domain to be pursued. Section \ref{Sec: Framework} will propose the conceptual framework for XRL and discuss how this integrates with human models of explainability. Section \ref{Sec:Review} will provide a review of current approaches to the initial stage of the framework, while section \ref{Sec:Opportunities} will identify future research opportunities for the advanced stages of the framework. Section \ref{Sec:Using} will discuss how the framework can be integrated into models of communication to better facilitate the development of \textit{Broad-XAI}. Finally, section \ref{Sec:Conclusion} will summarise the paper and its contributions.

\section{Explainable Artificial Intelligence}
\label{Sec: XAI}

\citeauthor{Harari2016} \cite{Harari2016} suggests that humans have always been a socially oriented species that have utilised their unique ability to articulate myths as integral to the social fabric. A myth is a story that aims to \textit{explain} historical events or natural/social phenomena \cite{MerriamWebster2020}, which helps guide future behaviour. There is no onus on the myth to be grounded in fact (e.g. religion, ideology, nationalism, money), simply that it provides an explanation for the world that allows us to operate socially --- if everyone else accepts the myth then large groups of people can be socially coordinated \cite{Harari2016}. The emphasis here is that explanation is fundamental to human social interaction and trust, and therefore, key to the social acceptance of artificial intelligent agents. However, while explanation has been studied by philosophers since Socrates, and over the last fifty years by psychologists and cognitive scientists, what it actually is, is still an open question \citep{Woodward2017Explanation}. As with the development of Artificial Intelligence, where research is hampered by people's poor understanding of intelligence, research into explainability is similarly restricted by a poor understanding of human explanation.

EXplainable Artificial Intelligence (XAI) is the general title given to the field of research aiming to generate explanations of AI systems that satisfies people's requirements in understanding and accepting the decisions made. There is a huge body of work providing a range of ways of interpreting black-box algorithms with mostly limited success. Various surveys have reviewed some of this work \cite{abdul2018trends, adadi2018peeking, dazeley2021levels}. \citeauthor{miller2017explainable} \cite{miller2017explainable}, however, argues that the majority of researchers make XAI systems that are specific to their area of AI and that the primary aim behind these systems is to debug --- rather than also considering the end-users' requirements. For instance, there are many explanation systems developed for image processing Convolutional Neural Networks (CNN) that universally focus on identifying areas of an image or the parts of the network that contributed the most to a particular result \cite{zhang2018visual}. \citeauthor{dazeley2021levels} \cite{dazeley2021levels} suggests that these `narrow' XAI approaches that only focus on the individual task at hand do not provide the details required by users of the ever increasing integrated intelligences currently appearing in the market. These emerging systems, such as autonomous cars, require \textit{Broad-XAI} approaches that merge the decision-making of several integrated systems into a coherent explanation \cite{dazeley2021levels}.

\begin{figure}
  \centering
  \includegraphics[trim={5.2cm 3.0cm 6.2cm 2.6cm},clip,width=20.0pc]{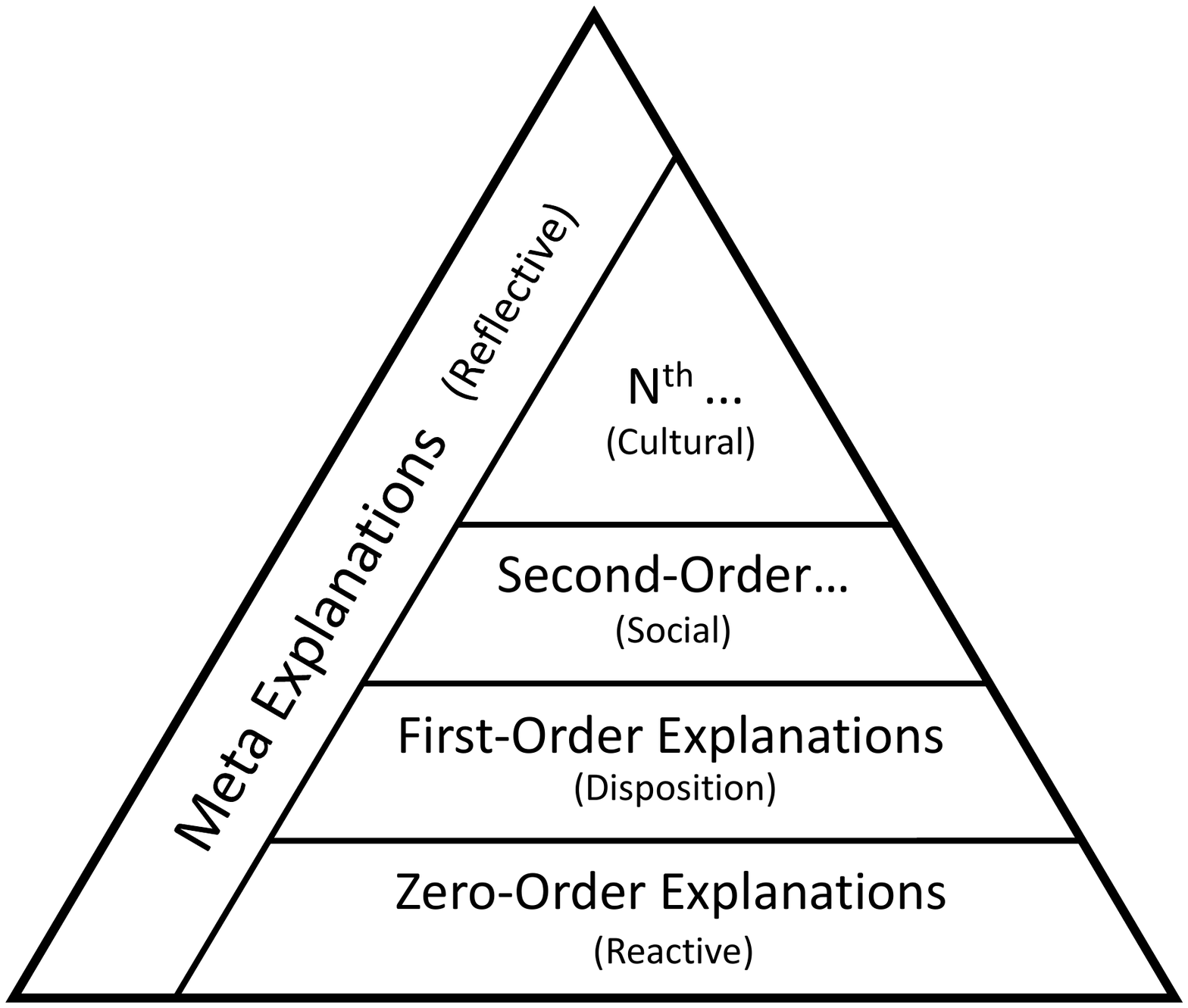}
  \caption{Levels of Explanation for XAI as proposed by \citeauthor{dazeley2021levels} \cite{dazeley2021levels}. Zero-order explanations cover an agent's immediate reaction to its environment and provides no indication of intentionality. First-order explanations justifies an agent's behaviour based on its current goal or objective, which is based on its current disposition. A Second-order explanation details an agent's behaviour in the context of the social situation it perceives, taking into account other actors in the environment. N\textsuperscript{th}-order explanations discuss how the agent has modified its behaviour on account of cultural expectations it believes are currently relevant. Meta explanations reflect on the processing process used in generating the explanation.}
  \label{fig:Levels}
\end{figure}

\citeauthor{dazeley2021levels} \cite{dazeley2021levels} suggest that most XAI research, which is often referred to as Interpretable Machine learning(IML), corresponds to Zero-order (Reaction) explanations --- where the zero refers to the absence of any explanation of the system's intentionality. Such approaches focus on explaining how the input just received was interpreted and how that input affected the resulting output. They argue that this foundational level is crucial to the development of Broad-XAI, but higher levels need to be developed for everyday users to accept decisions made by these systems. \citeauthor{dazeley2021levels} \cite{dazeley2021levels} suggests a set of levels, reproduced in Figure \ref{fig:Levels}, that build up an explanation based on the level of intentionality utilised when making the decision. For instance: First-order (Disposition) details an agent's intention such as its current goal or objective; Second-order (Social) justifies its behaviour based on a prediction of other actors' intentions; and, N\textsuperscript{th}-order (Cultural) provides an explanation of how it has modified its actions based on what it believes other actors' expectations are of its behaviour. Interestingly, there have been several attempts to develop approaches for these higher levels. \citeauthor{dazeley2021levels}'s \cite{dazeley2021levels} meta-survey identifies diverse subfields of XAI research, such as Explainable Agency \cite{langley2017explainable}, Goal-driven XAI \cite{anjomshoae2019explainable, anjomshoaeintelligible}, Memory-aware XAI \cite{kaptein2017role, rorty1978explaining, o1994explaining}, Socially-aware XAI \cite{hao2019emotion,mathews2019explainable, weitz2019you, sindlar2011programming}; Cultural-aware XAI \cite{kampik2019explaining, hellstrom2018understandable, wortham2017robot, dragan2013legibility}, Meta-explanation \cite{pitrat2006meta, galitsky2016formalizing, Galitsky2010ExplanationVM},  Utility-driven XAI \cite{ehsan2019automated, Ehsan2019OnDA, mclaughlin1988utility} and Deceptive Explanations \cite{person2019agents,van2014dynamics, sakama2015formal, sakama2014formal, sakama2011logical, sakama2010many, nguyen2011asp, zlotkin1991incomplete}.

All these sub fields, however, are focused on developing approaches for explaining that individual component of an explanation, whereas, Broad-XAI requires an integrated approach across all levels that affect an agent's decision. RL is a Machine Learning technique that potentially covers all these levels to some degree and offers a starting point for developing integrated explanations. However, currently research in this space is relatively limited. Hence, the aim in this paper is to present a conceptual framework of how RL can be used to provide explanations across all levels of explainability, and thereby, provide a foundation for the development of Broad-XAI.

\subsection{Explainable Reinforcement Learning: Temporal Explanations}
\label{Sec: XRL}

Most introductory texts on Machine Learning (ML) identify three subfields: Supervised, Unsupervised and Reinforcement Learning (RL) methods. RL is often identified as separate and distinct to other ML because it utilises a fundamentally different approach to learning. In RL an agent learns by interacting with an environment using trial-and-error learning. While trialling a sequence of actions it will occasionally receive feedback in the form of a positive or negative reward, which it will then attribute to those actions taken, reinforcing those behaviours to increase or decrease their selection in the future. 

\begin{figure}
  \centering
  \includegraphics[trim={1.8cm 5.8cm 4.5cm 6.5cm},clip,width=20pc]{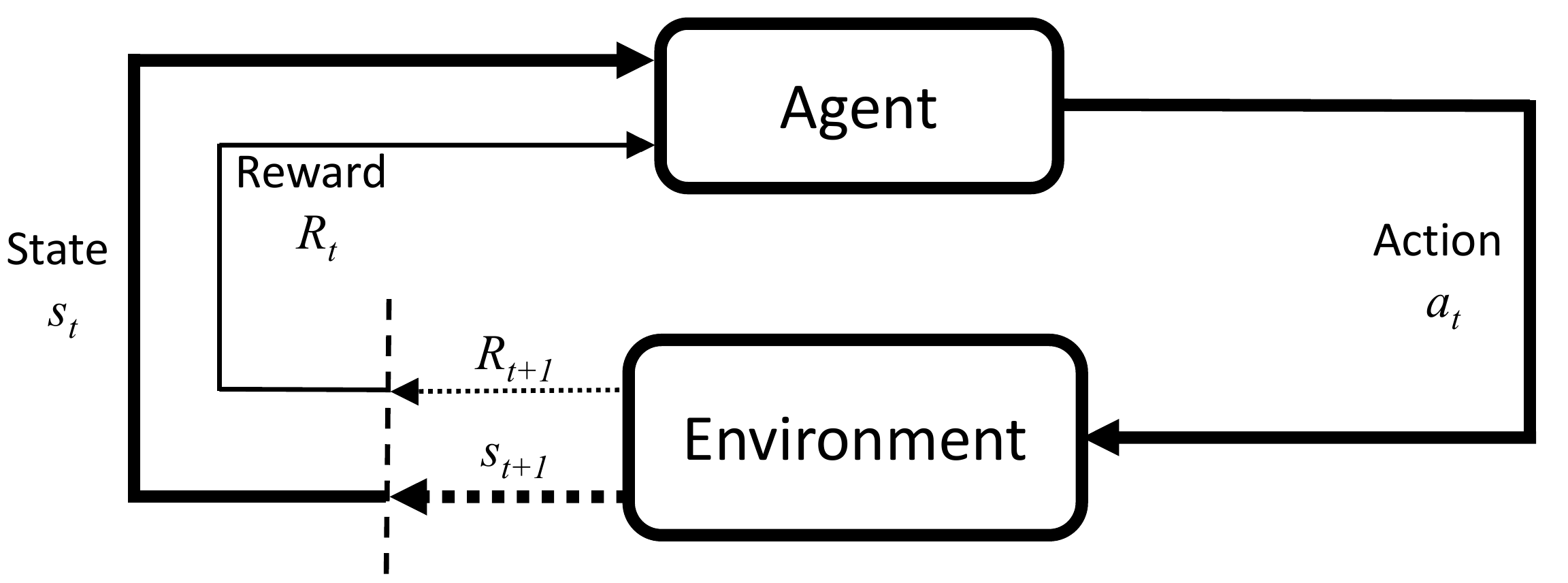} 
  \caption{Standard Reinforcement Learning model as presented by \cite{Sutton2018book}. An agent interacts with an environment through a series of discrete interactions. Each iteration it receives the current state of the environment and decides on an action to be taken. The next iteration it receives the new state and sometimes will also receive a numerical, positive or negative, reward which is used to train the previous choices.}
  \label{fig:RL}
\end{figure}

This has similarities to supervised learning, in that an agent learns a mapping from input (state) to output (action), but unlike supervised approaches the reward can be distributed temporally, as it may not receive the reward until many actions have been taken. Formally, as defined by \citeauthor{Sutton2018book} \cite{Sutton2018book} and shown in Figure \ref{fig:RL}, in the RL model the agent and environment interact through a series of discrete time steps, $t$. Each time step the agent receives a representation of the environment's current \textit{state}, $s_t \in \mathcal{S}$, where $\mathcal{S}$ is the set of all possible states. In a fully Markov Decision Process (MDP)\footnote{RL is also often applied in environments that are not fully Markov. For instance, Semi-MDPs require information from previous states to determine outcomes taken in future states. There are also significant work into Hidden, Partially Observable, Continuous-time, Multiobjective Markov processes etc.} the agent uses only this state information to select an \textit{action}, $a_t \in \mathcal{A}(s_t)$, where $\mathcal{A}(s_t)$ represents the set of all possible actions in state, $s_t$. In the subsequent time step, $t+1$, the agent receives a numerical \textit{reward}, $R_{t+1} \in \mathcal{R} \subset \mathbb{R}$, along with the new state, $s_{t+1}$. 

Essentially, an RL agent learns a mapping from each state to an action, which expresses the agent's behaviour. In model-based methods the agent optimises the trajectory of its behaviour to minimise cost, while value-based methods maximise the reward explicitly through a value-function. This mapping is commonly referred to as a \textit{policy}, and is denoted $\pi$, where $\pi(s, a)$ represents an individual mapping from state, $s$, to action, $a$\footnote{Methods not used for prediction rather than control may often learn a value per state rather than state/action pairs.}. 



There are numerous extensions to the basic RL approach that are frequently used in the literature. These are not the focus of this paper, but they do frequently present interesting information to the RL approach that allows for significantly improved explanations, and hence need some discussion. For instance, one difficulty with RL is as the state space grows so does the complexity of the agent's search for a solution. Hence, function approximation techniques such as Neural Networks are frequently utilised, giving rise to the field of Deep RL (DRL) \cite{mnih2015human, li2017deep, arulkumaran2017brief, duan2016benchmarking, hossain2019comprehensive, van2018deep}. Secondly, while most RL assigns a single goal to an agent, such as pick up the rubbish in the room and put it in the bin, there is substantial work in multi-goal RL. In such systems the agent not only must achieve its goal, but must also select the appropriate sub-goal to pursue \cite{barto2003recent, gebhardt2020hierarchical, zhou2020temporal, botvinick2012hierarchical,barto2004intrinsically, florensa2017automatic}. Finally, a goal represents the agent's ultimate objective, however, Multiobjective RL (MORL) assumes that there can often be other conflicting objectives that also need to be balanced with the primary objective \cite{roijers2013survey}. For instance an agent may have the goal to tidy the room and therefore its primary objective is to do this efficiently, however it may have a secondary objective to not damage any delicate things while accomplishing the primary objective \cite{vamplew2018human, vamplew2020potential}. 

RL, from an explanation point of view, is of particular interest as it is often regarded as differing to that of supervised learning approaches \cite{alharin2020reinforcement}. Supervised Learning techniques map each input to an output individually and so on their own the only explanation required is to identify the input components or the processes or processing stages that created the resulting classification. Each instance classified is regarded as a standalone instance and any \textit{local} explanation is inherently based on this fact. Additionally, classifiers may provide \textit{global} explanations that show how particular hyper-parameters or sets of training examples caused different outcomes for the classifier as a whole \cite{schwab2019cxplain,o2020generative,zhang2018equality}. These causal explanations are important for system developers or designers to understanding issues like training bias \cite{zhang2018equality} etc. However, supervised methods do not typically provide a mechanism for providing \textit{local} causal explanations that explain individual decisions or behaviours of a system for non-technical end-users.

RL based systems, however, have an implicit relationship between each instance. This is because the next state has only been visited because of the action taken in the previous state\footnote{Assuming a purely deterministic MDP. Clearly in many situations the new state only partially results from the previous state/action and sometimes other factors could also contribute to the resulting state, such as wheel slippage, other actors in the environment etc.}. This creates a temporal dependency between states, actions and subsequent states. These temporal dependencies, typically referred to as \textit{transitions} and denoted as $T(s_t, a, s_{t+1})$, provides an implied causation for that individual transition. A sequence of transitions, either when reflecting on past transitions or a prediction of future transitions, can potentially provide causal networks that can be used to explain a number of details such as why actions were chosen according to some long term goal \cite{cruz2020XRobot}. So, while an individual transition is similar to an individual classification in supervised methods the temporal sequence of transitions allows us to provide causal-based temporally-extended explanations. 

Additionally, supervised learning uses the learnt mapping to provide a classification or regression value with the aim of getting the \textit{`right'} answer, whereas, RL aims to maximise a reward signal, which symbolises the goal or objective of the agent. Many approaches to RL have been developed to identify sub-goals \cite{vattam2013breadth, Plappert2018MultiGoalRL, Caruana1997MultitaskLearning, Barto2003HierarchicalRL, Al2015hierarchicalRL, Teh2017MultitaskRL, Kaelbling93MultitaskLearning}, or may have alternative objectives that it can switch between, such as \cite{roijers2013survey, vamplew2017steering, vamplew2020potential}. These approaches mean the aim of the agent that guides its behaviour is not automatically going to be known to people affected by an agent operating in a human-agent shared environment. These approaches allow us to explain an agent's intentionality behind its behaviour, and thus, facilitate the provision of First-order explanations \cite{dazeley2021levels}. 

These fundamental differences between RL and supervised approaches to machine learning require us to think differently about explanation than simple interpretation --- the common approach in Machine Learning. Of interest is the ability to provide introspective, causal and contrastive explanations within a single platform. RL is an approach that allows us to potentially develop broad-XAI systems. The aim of the remainder of this paper is to develop and present a conceptual framework for the development of Broad-XAI utilising RL as the basic backbone. Within the context of this framework this paper will survey current attempts to provide explanations (section \ref{Sec:Review}) and discuss potential approaches, not yet attempted, that will promote further research and development in to Broad-XAI (section \ref{Sec:Opportunities}). 

\section{Conceptual Framework for Explainable Reinforcement learning (XRL)}
\label{Sec: Framework}

People interpret the world through explanations --- either by attributing explanations to others' behaviour or by explaining their own behaviour to themselves or others. When moving away from simply interpreting the decision-making process, as done by IML, developers need to consider how people tend to assign causes to behaviour \cite{heider1958psychology, jones1965acts, kelley1967attribution, kelley1973processes}. Attribution theory, based on \citeauthor{heider1958psychology}'s \cite{heider1958psychology} seminal work, attempts to understand the process to which people attribute causal explanations for events \cite{fiske1991social}. Most events are usually categorised as either being dispositional or situational. Dispositional attribution assigns the cause to the internal disposition of the person such as their personality, motives or beliefs. In contrast, situational attribution assigns cause outside the person's control such as accidents or external events. More recently, researchers have shown that people instead tend to attribute behaviour towards the person's intention, goal, motive or disposition \cite{malle1999people, malle2006mind, malle2000conceptual, kammrath2005incorporating}. 

\begin{figure}
  \centering
  \includegraphics[trim={1.4cm 2.2cm 1.2cm 2.8cm},clip,width=21pc]{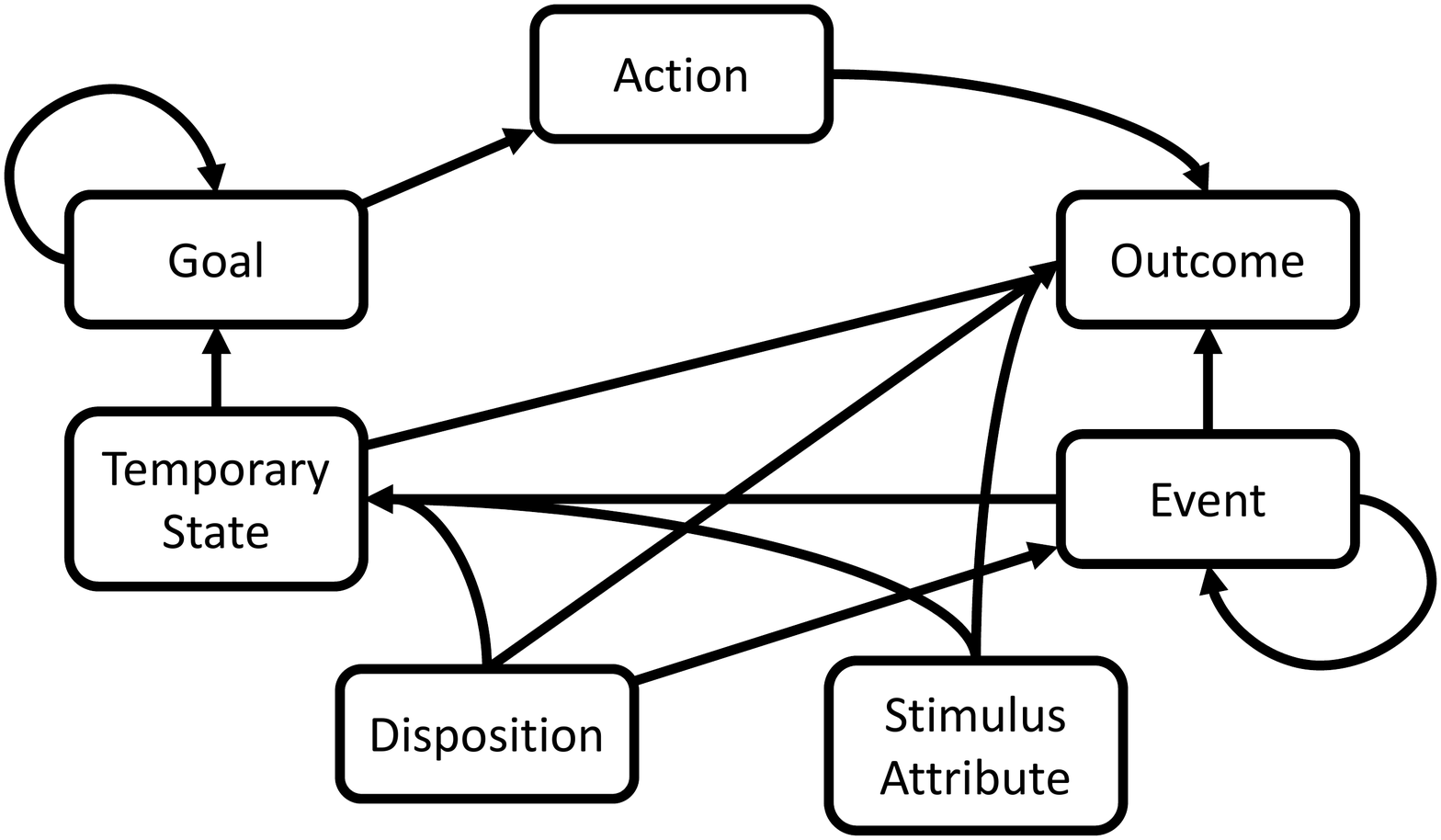}
  \caption{A reproduction of the Causal Explanation Network (CEN) model for human lay causal explanations as suggested by \citeauthor{bohm2015people} \cite{bohm2015people}. Each node represents a component used by people when explaining a person's behaviour, while the arcs between nodes indicate the causal links between these concepts when people provide an explanation.}
  \label{fig:CEN}
\end{figure}

Drawing on knowledge structures such as scripts, plans, goals, and themes suggested by \citeauthor{schank2013scripts} \cite{schank2013scripts}, \citeauthor{bohm2015people} \cite{bohm2015people} extended ideas in attribution theory to develop a Casual Explanation Network (CEN), Figure \ref{fig:CEN}, based on the actual explanations provided by people. This model emphasizes preconceptions about a causal relationship when providing explanations of behaviour. It builds on the idea that people will often want to explain others' behaviour not only in terms of why a particular behaviour occurred, but also what happened before to cause that behaviour and what is likely to happen in the future. \citeauthor{bohm2015people} \cite{bohm2015people} propose a taxonomy that classifies both behaviour and explanations and is built around the intentionality that lead to the behaviour.

The CEN, Figure \ref{fig:CEN} identifies seven categories that are relevant when considering the causal thinking about an actor's behaviour. This network is represented with a directed graph consisting of two sources and one sink. The end point, or sink, is the \textit{outcome}, which is the final result of any behaviours. These outcomes are a result of either a person's intentional goal-directed \textit{actions} or as a result of unintentional and uncontrolled \textit{events}, such as tripping over. A person's \textit{goal}, represents the future states that the person is striving for, which can be caused by higher-order goals. The goal can also be caused by the \textit{temporary state} or what can be thought of as their momentary disposition based on emotions, evaluations, mental states, motivational states, or bodily states (e.g. hunger, pain). This temporary state (momentary disposition) is in-turn affected by the person's personality traits or attitudes, which refers to as \textit{disposition}, that are the result of long term ingrained culturally-based behaviours. The temporary state can also be caused by \textit{stimulus attributes}, representing the features of the person or object that their behaviour was directed. For example, a person explaining the outcome of only passing an exam may state that it was too difficult (stimulus attribute) causing them to be upset (temporary state) so they altered their goal to make sure they at least passed. 

Figure \ref{fig:CEN} shows causal lines between these nodes indicating the causal directions provided in a person's explanations. These do not necessarily reflect the full and direct sequence of causes for outcomes, but they do represent the causal explanations that people \textit{typically} use \cite{bohm2015people}. For instance, if a person trips (event) they may explain that they are clumsy (disposition) and that fearing injury (temporary state), attempt to arrest their fall (goal), by reaching out their hand (action) resulting in scratches on their hand. When asked what happened to their hand, they may provide the full causal path or simply explain the shortened causal path indicating they had tripped. This allows the explainee to fill in the gaps with their own general understanding of probable causes. Similar choices are provided for causal paths between other nodes. In this way, an explanation does not always require the full causal path from event, stimulus attribute or disposition through goal and action. This approach elegantly agrees with \citeauthor{lombrozo2007simplicity}'s \cite{lombrozo2007simplicity} suggestion that an explanation should rely on as few causes (simple) as possible that covers the outcomes. 

The CEN's focus on causal behaviour being the basis of explanations of intentionality aligns with \citeauthor{dazeley2021levels}'s \cite{dazeley2021levels} suggested levels of explanation for XAI. These levels were built upon Animal Ethology's idea of explaining behaviour through levels of intentionality \cite{dorothy1990monkeys}. Furthermore, the taxonomy of causal behaviour suggested in the CEN aligns well with the operating paradigm of an RL agent, and therefore, its application to XRL would be useful in providing structure to the generation of causal explanations from an RL agent. This paper proposes to merge these ideas from \citeauthor{dazeley2021levels} \cite{dazeley2021levels} with the CEN, suggested by \citeauthor{bohm2015people} \cite{bohm2015people}, to form a framework, referred to as the Causal XRL Framework (CXF), and taxonomy for how XRL can generate causal explanations.

\begin{figure*}
  \centering
  \includegraphics[trim={0.4cm 2.2cm 0.4cm 2.8cm},clip,width=30pc]{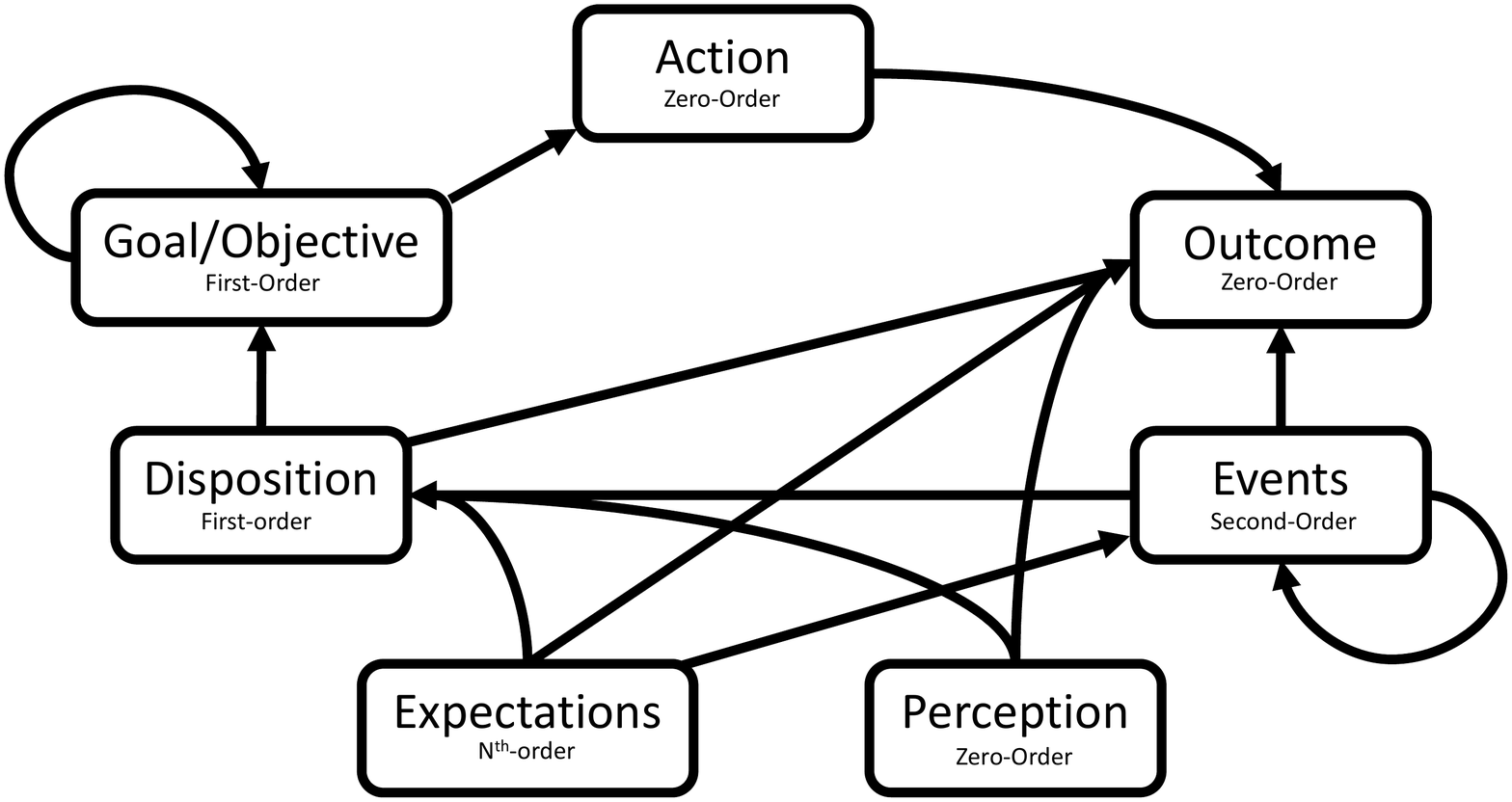}
  \caption{Conception Framework for Explainable Reinforcement Learning, referred to as the Causal XRL Framework (CXF) is based on the CEN given in Figure \ref{fig:CEN}. Each node represents a process used by an agent when deciding on its behaviour. Each arc, joining nodes, represent the causal relationships that should be utilised when generating an explanation of an agent's behaviour.}
  \label{fig:Framework}
\end{figure*}

Figure \ref{fig:Framework} is an adaptation of Figure \ref{fig:CEN} to facilitate the same causal pathways for explanation, but with categories aligned to RL and those indicated by \citeauthor{dazeley2021levels}'s \cite{dazeley2021levels} suggested levels of explanation. Included in this diagram is a mapping of XAI levels indicating the degree of intentionality that can be provided at each category of behaviour. This causal structure is intended to operate in a similar way to that suggested by \citeauthor{bohm2015people} \cite{bohm2015people}. An \textit{outcome}, represented by changes in the environment or the agent itself, is caused by either an intentional \textit{action} by the agent or by an unintended or uncontrolled sequence of events. These \textit{events} could be due to stochastic actions, such as wheel slippage or external actors. 

In RL an action is caused by an agent pursuing a particular \textit{goal} or \textit{objective}. This may be a single goal or a hierarchy of goals, each of which can be cycled through to generate the explanation of its behaviour. A goal may be aligned to a single objective or to multiple objective that must be balanced \cite{hayes2021practical}. The agent switch between these goals/objectives due to internal changes in priorities or progression in solving a larger goal. These internal changes are what \citeauthor{bohm2015people} \cite{bohm2015people} labels as temporary states, however this name could be confused with the perceived state of the RL agent, and hence, is avoided in CXF framework. \citeauthor{dazeley2021levels} \cite{dazeley2021levels} on the other hand refers to this same concept as disposition --- referring to an agent's internal disposition. Therefore, to align with \citeauthor{dazeley2021levels} \cite{dazeley2021levels} a \textit{disposition} in this sense is the same as a temporary states in the CEN model, and represents temporary internal motivations such as a change in parameter, simulated emotion or safety threshold being passed. 

Similarly, \citeauthor{bohm2015people} \cite{bohm2015people} CEN model referred to disposition as an overarching set of long term personality traits about how a person responds to situations. While there is no direct reference to responses to perceived cultural expectation, it is clear that disposition is the node where this would be best captured. As the temporal state node was renamed to be disposition the disposition node has also been renamed to align with \citeauthor{dazeley2021levels} \cite{dazeley2021levels} notion of cultural expectations. Therefore, in this model an \textit{expectation} refers to the ultimate aim of the agent to achieve what is expected of it. \citeauthor{dazeley2021levels} \cite{dazeley2021levels} suggests that expectations refer to a range of cultural conditions placed on an agent's operation. In essence, expectations in this framework are the same as dispositions in the CEN. Finally, an agent's current disposition, and therefore its goal/objective, action and ultimately the outcome, are caused by what is perceived by the agent. \textit{Perception} is both the literal state, but also the result of any feature extraction, inference placed over what is perceived, or belief state in a Partially Observable MDP (POMDP). 

Additionally, this framework is readily applicable to Multiagent Reinforcement Learning (MARL) domains \cite{chu2019multi}. For example, a MARL agent operating globally can simply use this framework directly with the understanding that the action space is a vector of actions that are similarly derived from its goals and higher-order influences. This aligns and extends current state-of-the-art explainable MARL \cite{kazhdan2020marleme}. However, a decentralised model presents a larger problem for the provision of explanations. A decentralised model requires agents to act independently of each other, and therefore, provide explanations of their behaviour independently. However, these agents require a sophisticated communication model between the agents to allow them to adjust their behaviour based on the other agents \cite{chu2019multi}. The CXF framework directly facilitates this MARL model. For example, when an agent changes its behaviour because of another agent's communication or action then the CXF model allows us to incorporate this behaviour as a causal event that potentially alters the agent's intrinsic disposition and goals. This approach allows for sophisticated models of explanation that incorporate teamwork directly into the causal framework. This decentralised model can be further extended to AI-Human collaborative teams \cite{lewis2018role,bethel2012discoveries} where we require an explanation of an agent's action in response to events caused by the human collaborators.

\begin{figure*}
  \centering
  \includegraphics[trim={2.6cm 11.8cm 1.9cm 2.8cm},clip,width=30pc]{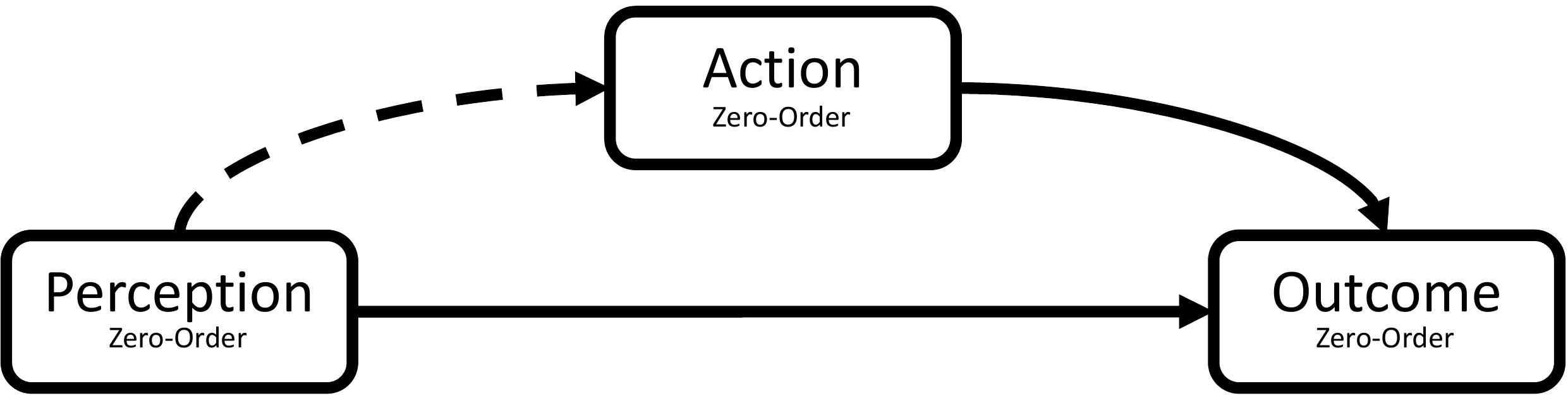}
  \caption{Simplified Conception Framework for Explainable Reinforcement Learning, referred to as the Simplified-CXF, representing causal explanations in traditional RL. This framework includes a causal link between perception and action that is not included in the full model. This link replaces several behavioural components representing deeper causal paths that are assumed to not be modelled or explained in this Simplified-CXF. Note, this simplified model only includes Zero-order explanations \cite{dazeley2021levels} and therefore does not include any explanation of intentionality.}
  \label{fig:Simplified Framework}
\end{figure*}

Ultimately, this framework is aimed at promoting future directions of research into explaining RL behaviour, but it also provides a lens for examining the current state of the art. The framework described in Figure \ref{fig:Framework} is beyond the majority of current XRL research. Hence, this paper also presents a Simplified Conceptual Framework, which captures the majority of current XRL work. The simplified framework, Figure \ref{fig:Simplified Framework}, shows the types of behaviours that can be explained when using a traditional approach to RL, as described in section \ref{Sec: XRL}. As can be seen, this model only includes behaviours caused by what is perceived and the actions taken by the agent. It can also be observed that these behaviours all align with Zero-order explanations \cite{dazeley2021levels}, and therefore, do not include any explanation of intentionality.  

In this simplified model it is assumed that an agent has a single preset goal and the objective is to maximise the reward in achieving that goal. In such a situation the goal is often known to the user or can be observed over-time through observation of behaviour \cite{huang2019enabling}. When utilising a predefined goal as its only objective its actions are directed toward achieving that goal. Its explanation of those actions are the target of that behaviour. \citeauthor{dazeley2021levels} \cite{dazeley2021levels} argues that there is no need to explain that the action is aimed at accomplish the goal in such a system. In situations where the goal itself is possible unknown to an end-user then the developers can incorporate details of the preset goal directly into any explanation of its behaviour. Equally, if the agent cannot alter its goal then no change to disposition or expectation can affect the goal being pursued. The aim of the \textit{Goal} node in Figure \ref{fig:Framework} is aimed at identifying how the current goal affected the action selected and why that is the current goal based on the agents current dispositions or expectation. Hence, the simplified model has no need to include causal explanations of these higher-level intentions. Similarly, the general RL model makes no attempt to model events outside its control making explanations of these also irrelevant. With the removal of goal/objectives, dispositions, expectations and events an RL agent cannot utilise those causal paths, therefore, the simplified framework must include a causal path from perception to action skipping those behaviours included in the full framework. Because this causal path is not part of the full framework this is included only as a dotted line.

\section{Simplified Framework: Reviewing Explainable Reinforcement Learning}
\label{Sec:Review}

The term, eXplainable Reinforcement learning (XRL) only appeared in research publications recently and is often published as Interpretable Machine Learning (IML). However, the aim of this paper is to show that the idea of explaining the behaviour of an RL agent, while sometimes related, is often quite distinct and separate to traditional IML; provides opportunities for deeper explanations to provide user trust and acceptance; already has a substantial body of research; and, still has significant avenues for future work. This section represents the second substantive component of this paper, which will review current work and discuss opportunities for future research. Rather than using a traditional taxonomy of approaches, it will review the literature in light of the Simplified-CXF discussed in section \ref{Sec: Framework}.

The following subsections will discuss each of the processes used by an agent to influence its choice of behaviour. This will include a discussion of the possible types of causal explanations that each process can contribute. Finally, for each type of causal explanation pathway, this paper will both discuss current approaches to explaining that causal link, as well as suggest additional approaches that could be utilised. In discussing these points it will start with the nodes represented in the Simplified-CXF and discuss the opportunities available in the more advanced components of the full CXF in section \ref{Sec:Opportunities}. The first subsection \ref{Sec:PerceptionsSimple} will discuss explanations of what the agent has perceived and how that perception has affected the actions and outcomes. Subsection \ref{Sec:Actions} will discuss explanations based on why actions are selected and how they caused the resulting outcomes. 


\subsection{Explanation of Perceptions}
\label{Sec:PerceptionsSimple}

At its fundamental level, an RL algorithm is learning to do two things: receive information about the environment and use this to decide on an action to make in response. These two fundamental operations of an RL system represent the first two types of XRL discussed in this paper. The fundamental nature of these operations are also indicated in Figure \ref{fig:Simplified Framework}, with them being recognised as providing Zero-order explanations \cite{dazeley2021levels}. That is, these operations represent a purely reactionary level of processing with zero intentionality. The first of these operations is to perceive the environment, which represents a significant amount of research in XRL. This section will briefly overview this class of XRL and discuss some example approaches. As identified by the simplified conceptual framework, Figure \ref{fig:Simplified Framework}, the perceptual stage, not only explains what it has perceived, but also how that perception resulted in the action taken and the outcome observed. Therefore, explanations of an agent’s perception aim to detail one or more of the following:

\begin{enumerate}
    \item \textit{Perception:} what did the agent perceive as the current environment?
    \item \textit{Introspective:} how the perceived state contributed to the action being selected?
    \item \textit{Contrastive:} why didn't the perceive state cause some other action to be selected?
    \item \textit{Counterfactual:} what changes in perception would be required to cause an alternative action to be selected?
    \item \textit{Influenced:} how did the perceived state affect the outcome?
\end{enumerate}


In the simple discrete RL situation each state or state/action pair can be represented with a mapping directly to the preferred action. However, in most realistic problems the state dimensionality for a direct mapping is too complex or continuous preventing a direct mapping. Instead, one of several approaches can be used such as function approximation \cite{busoniu2010reinforcement}, hierarchical representations \cite{botvinick2009hierarchically, botvinick2012hierarchical, barto2003recent}, state aggregation \cite{singh1995reinforcement, hutter2014extreme}, relational methods \cite{van2005survey} or options \cite{sutton1999between, bacon2017option}. 

To perform these approximations RL researchers generally utilise a range of traditional supervised learning approaches. For instance, the utilisation of Deep Neural Networks (DNN) is so common that a separate branch of research, known as Deep RL (DRL), has emerged, which now represent approximately 32\% of RL papers published in 2019\footnote{At time of writing, using Google Scholar, the number of papers with a title including the phrase "Deep Reinforcement Learning" was 1710 and the number of "Reinforcement Learning" titled papers was 5310. This is a crude estimation and almost certainly lower than the true percentage as many researchers assume DRL when discussing RL.}. DRL methods utilise a DNN to map large state spaces to Q-values (regression) or directly to actions (classification) \cite{mirowski2016learning}. In many cases the supervised learning model used requires some level of adaptation to handle the temporal aspects of RL. For instance, DRL methods frequently utilise various forms of experience replay to improve convergence \cite{schaul2015prioritized}. However, regardless of the learning process, the perception of the environment at any single moment is essentially the same process used in the supervised version.

\subsubsection{XRL-Perception with Interpretable Machine Learning (IML)}
\label{Sec:IML}

Reliance on traditional supervised learning for function approximation means that XRL-Perception is essentially the process of interpreting the function used to model the state. Therefore, XRL-Perception is closely aligned with \textit{Interpretable Machine Learning} (IML) methods \cite{abdul2018trends, adadi2018peeking, doshi2017towards, molnar2019interpretable, gilpin2018explaining}. IML is a well-established field with substantial work already having been done. The aim of this paper is not to resurvey IML work in detail - except to discuss how this work can be related to XRL specifically. According to \citeauthor{molnar2019interpretable} \cite{molnar2019interpretable} there are several approaches to interpreting machine learning models, as shown in Figure \ref{fig:IML}. This suggests that IML typically produces one or more of the following types of interpretation:

\begin{figure}
  \centering
  \includegraphics[trim={1.2cm 4.2cm 0.3cm 2.5cm},clip,width=20pc]{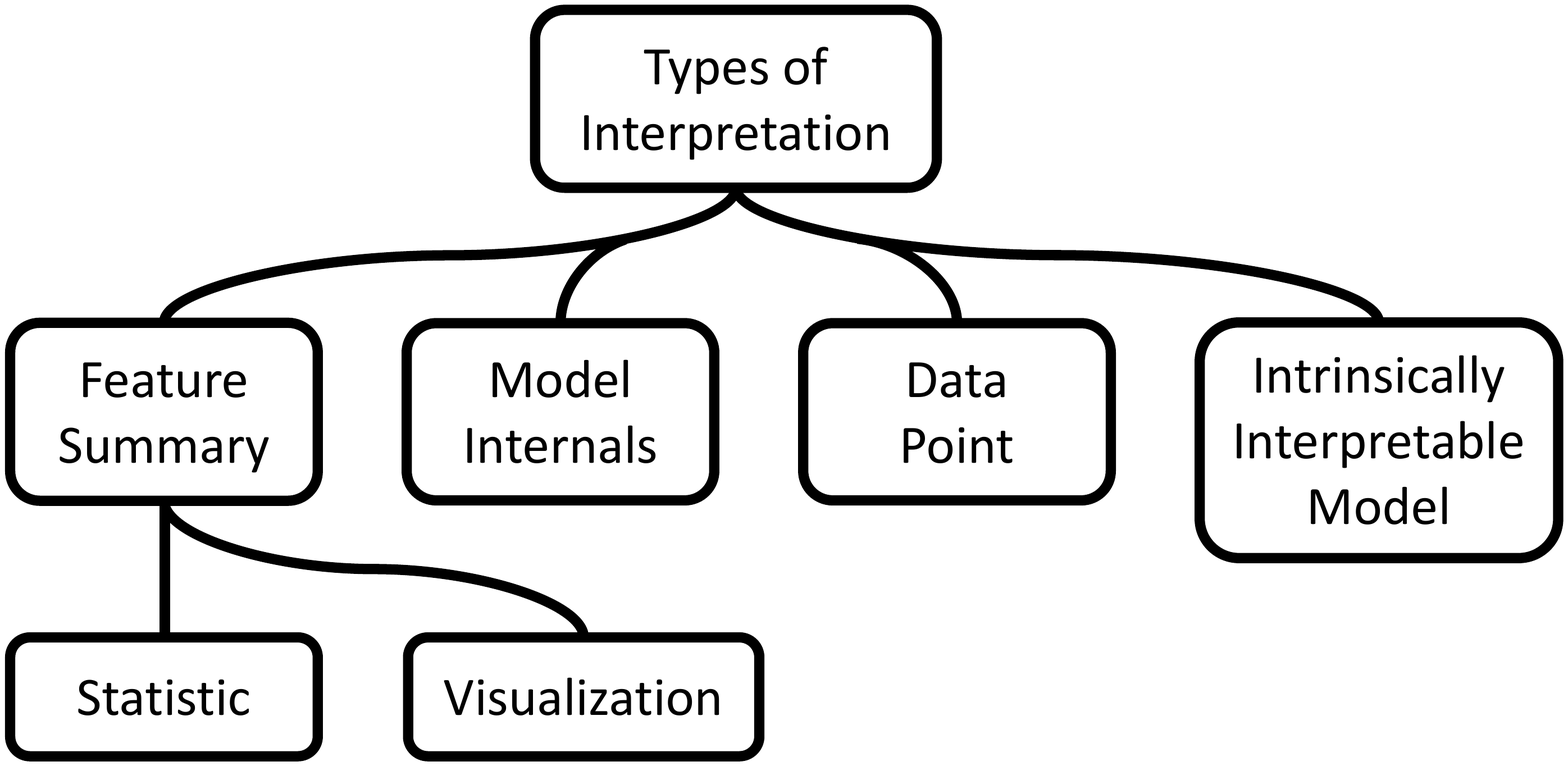}
  \caption{Types of Interpretation that can be generated from an Interpretable Machine Learning model. This diagram is derived from the taxonomy described by \citeauthor{molnar2019interpretable} \cite{molnar2019interpretable}.}
  \label{fig:IML}
\end{figure}

\begin{itemize}
    \item a \textit{feature summary}, using \textit{statistics} or \textit{visualisations}, showing the features and their relationships that were of most importance when reaching the outcome.
    \item a representation of the \textit{internal model's} operation, such as the rules or neurons that fired, or pathways through the evaluation process that were followed.
    \item through the identification of similar or related \textit{data points}, such as an image from the same class.
    \item through the construction of a secondary \textit{intrinsically interpretable model}, which may then use one of the above methods to provide an interpretation. 
\end{itemize}

Deep learning methods for IML tend to focus on visualisations of features found in the input (feature summaries) and neuron/layer activity (internal models); with some examples of using specifically designed neural networks for the provision of interpretations --- see \citeauthor{gilpin2018explaining} \cite{gilpin2018explaining} for a detailed discussion. Regardless of the approach used to interpret these models they can all be utilised to provide an interpretation of the perception of current state of an RL model. 

\subsubsection{Introspective XRL-Perception}
\label{Sec:Introspective XRL Perception}

Due to this alignment of perception and traditional IML there has been limited research specifically on perception in the context of XRL \cite{heuillet2020explainability, puiutta2020explainable}. However, there are two primary issues that make perception in XRL distinct from traditional IML. The first is that spatially similar states may often still require different control rules making generalising difficult. This contrasts with most traditional supervised approaches that can afford local generalisation. Secondly, the perceptually similar states may in fact be significantly temporally separated \cite{zahavy2016graying}. This problem is attributed to why pooling layers are often absent in many DRL approaches as these are used to identify local generalisable patterns \cite{mnih2015human, levine2016end, andrulis2020domain}. Therefore, research into XRL-perception has largely focused on providing explanations that can help developers to better understand the learning process, improve interpretation of policy, and for debugging/parameter tuning \cite{zahavy2016graying}. 

One approach used by both \citeauthor{mnih2015human} \cite{mnih2015human} and \citeauthor{zahavy2016graying} \cite{zahavy2016graying} employed t-Distributed Stochastic Neighbor Embedding (t-SNE) on recorded neural activations \cite{mnih2015human, zahavy2016graying} to identify and visualise the similarity of states. \citeauthor{zahavy2016graying} \cite{zahavy2016graying} also displayed hand crafted policy features over the low dimensional t-SNE to better describe what each sub-manifold represents. A second approach used by \citeauthor{wang2015dueling} \cite{wang2015dueling} and \citeauthor{zahavy2016graying} \cite{zahavy2016graying} was to use Jacobian Saliency Maps \cite{simonyan2014very} to better analyse how different features affect the network. \citeauthor{shi2020self} \cite{shi2020self} uses a self-supervised interpretable network (SSINet) to locate causal features most used by an agent in its action selection. 

These approaches are complex to understand and do not easily provide a reasonable explanation to a non-expert user. Saliency maps provide a reasonable level of understandability when using image-based state spaces but the Jacobian approach, borrowed from IML, can provide poor results as they have no relationship with the physical meaning of entities in the image. This problem can be exacerbated in an RL agent due to the spatial similarity of states. \citeauthor{greydanus2017visualizing} \cite{greydanus2017visualizing} improved this approach by utilising the unique dual use of networks in the Asynchronous Advantage Actor-Critic (A3C) algorithm to separately represent both the critic's value assignment and the actor's actions.  \citeauthor{greydanus2017visualizing} \cite{greydanus2017visualizing} then used these more accurate maps to visualise an agent's perception over time during the training process. This approach provides an important example for detecting features and identifying which features caused the agent to take a particular action, and separately, which ones were associated with particular outcomes, such as the highest rewards. 


\citeauthor{verma2018programmatically} \cite{verma2018programmatically} presented a unique approach to performing introspection of an RL agent's perception by altering the RL framework itself. This work introduced the Programmatically Interpretable RL (PIRL) approach, where policies are initially learnt using DRL. This network is then used to direct a search over programmatic policies using Neurally Directed Program Synthesis(NDPS). During this repeated search process, a set of interesting perception patterns are maintained that minimise the distance between the DRL and NDPS (oracle) models. The completed oracle can then be inspected to identify causal links between feature vectors and actions taken and/or outputs.

\subsubsection{Results of XRL-Perception}
\label{Sec:Results of XRL-Perception}

Perceiving the state is of particular interest to developers when validating a system's operation. Reassuring a non-expert user that the important features being used are also of importance, provided this is combined with the resulting effect of what was perceived. Simply informing the user of the action and the resulting change in the environment is implied in the previously discussed approaches as these are generally easily observed and do not require an explanation. However, the ability of a system to provide either contrastive or counterfactual explanations can be very valuable to a non-expert user and not easily observable from the agent's behaviour. Such explanation facilities, not only want to be able to identify the features that led to the selected action, but also suggest why another action was not selected (contrastive), or which features were needed to be observed to cause a different action/outcome (counterfactual). 

Conceptually counterfactual thinking and contrastive explanations are viewed as very different concepts. However, they are really just different views of the same predictive mechanism \cite{robeer2018contrastive}. A counterfactual focuses on a prediction of what would happen under different initial circumstances, whereas a contrastive explanation details what change was needed to get a particular outcome. A counterfactual can be derived by providing a case study, or example fictitious state (sometimes referred to as a `distractor' state or image), and observing the result. The real outcome, along with the fictitious outcome, can then be compared to provide the counterfactual explanation \cite{chang2018explaining, goyal2019counterfactual, atrey2019exploratory}. 

Contrastive explanations, however, are not as simple because there is no specific start state, but instead have a specific result that is of interest. The approaches in the last section cannot readily provide such explanations. For example, generating contrastive or counterfactual explanations requires us to identify the features that are missing from the input space. One approach is to present multiple distractors and find the closest to the required conclusion \cite{goyal2019counterfactual}. This, however, is highly computational, and impossible when there are infinite possible distractors, such as continuous state problems. Recent methods for generating missing features, such as the Contrastive Explanation Method (CEM) \cite{dhurandhar2018explanations, gu2018understanding}, have been proposed. These systems effectively identify absent pixels using a perturbation variable \cite{dhurandhar2018explanations} or through Contrastive Layer-wise Relevance Propagation (CLRP) \cite{gu2018understanding}.

In RL, however, there is a temporal relationship between states and the outcomes that can be used to map a sequence of changes over time. This creates additional possibilities for providing contrastive explanations, and thereby, through extension counterfactual explanations as well. One approach is to identify those states that are critical to a human understanding the result of an agent, such as the \citeauthor{huang2017leveraging} \cite{huang2017leveraging} utilisation of DBSCAN \cite{ester1996density} to identify such states. An alternative approach used recently, especially in non-image-based inputs was that of \citeauthor{hayes2017improving} \cite{hayes2017improving}. This work uses hand crafted state features specifically identified for being semantically meaningful to humans. This vector of features are used to generate a list of predicates that could be searched to identify subsets of actions commonly associated. Therefore, this approach could explain why an action was selected in terms of features perceived by the agent. This approach is not, however, readily usable in large state spaces or where hand crafted features could not be provided. There is potential in using an agent's perception to generate contrastive and counterfactual explanation, however, currently XRL researchers do not appear to have pursued this approach significantly. Instead, XRL focus has been around explaining the choice of actions and performing causal analysis of those choices. These approaches will be further discussed in the following section \ref{Sec:Actions}. 

\subsection{Explanation of Actions}
\label{Sec:Actions}

While the provision of explanations of an agent's perception is interesting, and in many cases required by the explainee, such explanations are not particularly unique to RL. In fact, in reviewing the literature above, very little referred to RL specifically. As discussed in section \ref{Sec: XRL}, the reason XRL is different to IML approaches is due to the temporal nature of RL. This temporality is evident when considering how an action taken by an agent affects the outcome. These explanations are inherently \textit{temporal explanations} as they detail a prediction of the expected future efficacy of an action. Temporal explanations detail relations between temporal constraints, such as delays between causes and effects and were first investigated in temporal abductive reasoning \cite{brusoni1997efficient} and recommendation systems \cite{bresina2006explanations, bharadhwaj2018explanations}. The CXF, Figure \ref{fig:Framework}, and simplified CXF, Figure \ref{fig:Simplified Framework}, indicate that an explanation can include why an agent took particular actions and how those actions caused particular results. 
Therefore, explanations of an agent's actions aim to detail either: 

\begin{enumerate}
    \item \textit{Introspective:} why was an action chosen?
    \item \textit{Contrastive:} why wasn't another action chosen?
    \item \textit{Influenced:} how the action taken affected the outcome?
    \item \textit{Counterfactual:} what prior behaviour would have resulted in a particular alternative action being selected?
\end{enumerate}

The first point addresses an explainee's requirement to understand the choice of action and why the agent predicts it is a better choice than the alternatives. This can be presented in one of two forms, either: providing a visual representation of the path; or, by stating how the action leads to the eventual aim. For example, imagine an agent takes an action a user wants justified. It could present a map showing where the agent is currently located and the path it plans to follow, where the user can see the selected action follows this path. They could also be shown the best path should an alternative action be taken. This approach is of course regularly used in navigation recommendation systems such as Google Maps. Non-navigation in discrete tasks can also use this approach by representing the MDP as a graph using nodes and arcs to represent concepts the explainee will understand. An alternative approach is to state that the agent has selected a particular action because it has a measurably better result of a desirable quality as defined by the reward function, such as a higher chance of success, reduced cost, safer, smoother, etc. In either case the agent is being asked to make a prediction about both its future behaviour and how it expects the environment to respond.

\subsubsection{Model-based XRL-Behaviour}
\label{Sec:Model-basedApproaches}

Early research into explaining why an action is preferred when accomplishing a particular task can be traced back to some of the earliest work in explaining the reasoning of expert systems \cite{shortliffe1975model, davis1977production, swartout1983xplain, chandrasekaran1988explanation}. An expert system generates a conclusion through a series of inferences. These inferences represent a sequence of reasoning steps that can be considered actions during a problem-solving process. Explaining these involved providing either a rule trace of the inferences/actions taken or a trace of key, previously identified, decision points. These early ideas were later extended in domains such as Bayesian Networks (BN) \cite{lacave2002review}, where explanations were generated from the relations between variables \cite{druzdzel1996explanation,renooij1998decision} or through visual representations of relations between nodes \cite{lacave2000graphical}. Decision Networks or influence diagrams further extended BNs through the incorporation of utility nodes. These models help the decision process by selecting the path with the maximum utility, where explanations have been generated by reducing the optimal decision table \cite{Bielza2004optimal}.


An MDP, as used in RL, can be considered to be a \textit{dynamic} decision network \cite{elizalde2007mdp, elizalde2009generating}. Similar approaches have been applied in deterministic or decision theoretic planning \cite{hanheide2017robot} because these focus on the entire decision path followed, and therefore, rely on a model of the environment. XAI-Planning (XAIP) approaches are well placed to provide explanations for planning tasks with MDPs. \citeauthor{fox2017explainable} \cite{fox2017explainable} provides a roadmap for the development of XAIP. These methods have a model of the environment in which they operate, and therefore, are inherently more transparent and understandable as they can use their model directly in their explanations. These approaches allow a more direct utilisation of the historical BN and DN methods. For instance, \citeauthor{krarup2019model} \cite{krarup2019model} uses waypoints for explanation, where this use of an execution-trace is a similar approach to that of rule traces and tracing nodes through a BN or DN. Similar approaches of generating explanations from actions using a model can be seen in other recent research \cite{dodson2011natural, elizalde2008policy, kasenberg2019generating, kasenberg2019engaging, chen2020towards, hoffmann2019explainable, chakraborti2017visualizations, gopalakrishnan2019tge, chakraborti2019planning, bongartz2018explaining, wang2016impact, seegebarth2012making, hein2018interpretable}. \citeauthor{fox2017explainable} \cite{fox2017explainable} identifies several questions that XAIP can answer. Ignoring questions regarding if and when to replan, which are specific to XAIP, these questions align with the previously mentioned aims for explaining actions. While planning approaches are not the focus of this paper, \citeauthor{chakraborti2020emerging} \cite{chakraborti2020emerging} provides an extensive and recent survey of XAIP identifying the recent growth.

\subsubsection{Introspective XRL-Behaviour}
\label{Sec:Introspective XRL-Behaviour}

A direct adaptation of the BN and DN approaches are not as evident in value-based RL. \citeauthor{cruz2019XRL} \cite{cruz2019XRL}, \citeauthor{hayes2017improving} \cite{hayes2017improving}, and \citeauthor{lee2019complementary} \cite{lee2019complementary} could be considered as attempts to do this by essentially developing a model of the environment during exploration. The models built can then be used to generate an explanation, such as a prediction of the likelihood of reaching a goal, and how long until it was reached, from each state/action pair. \citeauthor{hayes2017improving} \cite{hayes2017improving} learns its model entirely separately from the agent, while \citeauthor{cruz2019XRL} \cite{cruz2019XRL} build the model internally. The approaches are inherently still RL as the model is not used for planning purposes and the agent still learns entirely from experience. However, these approaches of building a model of the environment presents allow an RL agent to present a similar level and range of transparency exhibited by the model-based approaches.  

These learnt-model based approaches can also be used to provide users with an overview of the model through Policy Summarization or similar approaches \cite{lage2019toward, amir2018agent, amir2018highlights, huang2019enabling, lage2019exploring, sequeira2020interestingness, huang2018establishing}. These global explanation approaches learn key state/action pairs that globally  characterize  the  agent’s  behaviour. Using Inverse RL techniques, a policy can be inferred, and a summary formed from multiple examples of agent behaviour. The intuition is that policy summaries, like waypoints, can help people generalise and anticipate agent behaviour \cite{lage2019toward}. Another approach is to abstract away from low level decision and provide explanations from this higher level. \citeauthor{beyret2019dot} \cite{beyret2019dot} used Hierarchical RL to perform these layered abstractions and recognised their applicability to providing explanations and \citeauthor{acharya2020explaining}\cite{acharya2020explaining} used a decision tree classifier to learn which state features were most likely to predict particular behaviours. 


Ultimately, without a model, value-based approaches are hampered in their ability to explain an action in terms of the eventual aim. While people may think their aim is to achieve a goal, it is in fact only to maximise the long-term average reward. \citeauthor{cruz2020XRobot} \cite{cruz2020XRobot} extended their learnt model approach \cite{cruz2019XRL} to provide the same explanations without requiring the memory overhead of learning a model, thereby providing the ability to provide these explanations in larger environments, including those requiring deep learning based function approximation. To do this \citeauthor{cruz2020XRobot} \cite{cruz2020XRobot} proposed two approaches: learning-based and introspection-based. The first approach was to directly learn a probability value $P$ during training, while the second approach, referred to as introspection-based, was to infer the value directly from the agent's $Q$-value using a numerical transformation. These approaches allow an agent to explain why one action is preferred over another in terms of outcomes in a similar way as XAIP approaches. 

What is interesting about these approaches is that rather than learning a model they use introspection of available information to provide explanations. \textit{Introspection} is the utilisation of internal data for explanation as opposed to external frameworks that explain through observation. This introspective approach has also been utilised by \citeauthor{sequeira2020interestingness} \cite{sequeira2020interestingness}, which actively builds a database of historical interactions, allowing for simple information like, observations, actions and transitions; along with inferred probabilities such as the prediction error. While this work as presented is not built to specifically answer questions, it does provide details that can provide additional analytics to the user and could easily utilise these statistics to provide such answers. This work has since been extended to provide short video highlights of key interactions \cite{sequeira2020interestingness}.

\subsubsection{Results of XRL-Behaviour}
\label{Sec:Results of XRL-Behaviour}

When providing an explanation using the above techniques the system can simply state the reason the action selected is a good choice for achieving its goal. This, however, will often result in a relatively meaningless explanation that it chose the best, fastest, cheapest etc, depending on the choice of reward. Instead, as discussed in section \ref{Sec:Results of XRL-Perception}, an explanation aiming to improve trust and acceptance would ideally be presented in contrast to an alternative action. These \textit{contrastive explanations} are presented as \textit{fact} and \textit{foil} \cite{lipton1990contrastive, miller2018contrastive}, where the same fact, the action selected, can have multiple foils, any one of the actions not selected. Explaining contrastive and counterfactual of XRL-Behaviour involves comparing outcomes from alternative transitions paths through the MDP. 

The most common approach to providing these explanations is to develop a model of the agent's behaviour using a separate observer that learns the agent's behaviour. There have been several generic explanation facilities that can perform this task, such as \citeauthor{pocius2019strategic} \cite{pocius2019strategic}, which extends Local Interpretable Model-Agnostic Explanations (LIME) \cite{ribeiro2016should} and can provide contrastive explanations of any type of agents' behaviour --- not solely an RL agent. These generic explanation facilities can predict behaviour, but do not explain the agent's internal reasoning for its behaviour.

Extending \citeauthor{hayes2017improving} \cite{hayes2017improving}, \citeauthor{van2018contrastive} \cite{van2018contrastive} provides contrastive explanations based on the result of transitions. The approach uses a provided model of the transition network but acknowledges this can be learnt through the observation of behaviour, by translating state features and actions to a predefined domain specific ontology. The system then compares a user selected foil to the taken actions to provide explanations on outcome differences. \citeauthor{cashmore2019towards} \cite{cashmore2019towards} provides a generic planning wrapper that builds on \citeauthor{fox2017explainable}'s \cite{fox2017explainable} roadmap for XAIP, to provide these contrastive explanations for known MDPs as a service. Rather than an \textit{a priori} model \citeauthor{madumal2019explainable} \cite{madumal2019explainable} used a learnt model to extensively study the generation of both contrastive and counterfactual explanations for explaining recommendations in the game of Starcraft II \cite{vinyals2017starcraft}. \citeauthor{madumal2019explainable} \cite{madumal2019explainable} learns a Structural Causal Model (SCM) during training and analyses this model to understand how states led to different outcomes. 

To investigate the ability to provide a value-based approach, \citeauthor{cruz2020XRobot} \cite{cruz2020XRobot} illustrates that contrastive explanations on the likely success or failure of actions and the time to a result can be provided by an agent using the introspection-based approach to transforming the Q-values directly. \citeauthor{khan2009minimal} \cite{khan2009minimal} developed an approach to generate explanations for why a recommendation has been provided to a user, called a Minimal Sufficient Explanation (MSE). In this approach, a recommendation equates to an action and the approach tries to explain why that action is regarded as optimal. It takes one step beyond simply saying the action selected has the highest Q-value and thus is the optimal action, and instead provides reasons according to templated justifications about frequency of expected future rewards. 

Two possible approaches to providing contrastive explanations is through the utilisation of either reward decomposition \cite{juozapaitis2019explainable} or multi-objective Reinforcement Learning (MORL) \cite{roijers2013survey, vamplew2011empirical, vamplew2018human}. Reward Decomposition separates each of the different rewards into semantically meaningful reward types allowing actions to be explained in terms of trade-offs between the separate rewards \cite{juozapaitis2019explainable}. 

One avenue to providing contrastive explanation that has only recently been attempted is through the utilisation of multiobjective RL (MORL) \cite{roijers2013survey,vamplew2011empirical,vamplew2018human}. MORL approaches maintain a vector of Q-values for each reward and at any given time there may be several Pareto-optimal policies offering different trade-offs between the objectives. Such approaches, such as reward decomposition, allow an agent to compare the known results of these policies that aligned with different actions. \citeauthor{sukkerd2018toward} \cite{sukkerd2018toward} and its extension \citeauthor{sukkerd2020tradeofffocused} \cite{sukkerd2020tradeofffocused} along with work by \cite{juozapaitis2019explainable} are the first papers to directly pursue this approach to contrastive explanation. This model-based approach generates quality-attribute-based contrastive explanations to compare actions against alternative objectives. In value-based RL there is also one known attempt to use multiple objectives using reward decomposition\footnote{The authors do not identify their work as MORL, and strictly speaking decomposed rewards are not necessarily conflicting (a generally accepted component of an MORL problem), but the approach uses the same principles that would underpin an explainable MORL approach.} \cite{erwig2018explaining}. This approach is performing RL in an Adaptive Based Programming formalism that allows annotations of decision points with ontological information for explanation. Currently there is significant opportunity to pursue explainable MORL approaches for contrastive and counterfactual explanations.  

The above approaches assume there is only one foil, alternative action, or that the user knows which foil they want the agent to compare with the selected action. However, this can be tedious, difficult or sometimes impossible for the user to provide. For instance, in an autonomous car it is not practical to go through all alternative angles a steering wheel could have been turned to observe alternative results. Deriving the foil from the context is part of the explanation facility's task, but apart from some attempts in IML \cite{van2018contrastive, sokol2020one, robeer2018contrastive, rathi2019generating}, has not been widely discussed in the context of RL. While not the focus of their research some work in dynamic programming, such as \citeauthor{erwig2020explanations} \cite{erwig2020explanations}, have found that the context for contrastive explanations could be anticipated by identifying principal and minor categories and using these to anticipate user questions through value decomposition. As yet, foil prediction does not appear to have been transferred to value-based RL.


\section{Full Framework: Opportunities for Explainable Reinforcement Learning}
\label{Sec:Opportunities}

Explaining perception, action and the causal outcomes of each, discussed above, represent the majority of current XRL research. These explanation facilities are important but focus primarily on providing debugging style explanations for developers \cite{miller2017explainable, miller2017explanationBook}. \citeauthor{dazeley2021levels} \cite{dazeley2021levels} argued that this represents only a zero-order, or reactionary level, of explanation and does not provide the broad-XAI required to develop user trust and acceptance. While there is still plenty of scope for interesting advances in the above simplified-CXF, this paper suggests there are significant possibilities for higher-level explanations built on an RL foundation. This section will discuss each of the remaining components of the full framework and how existing extensions to RL can be utilised to provide Broad-XAI facilities in XRL.

\subsection{Explanation of Goals}
\label{Sec:Goals}

Explaining an agent's goal and how it caused the selected action has been recognised as a potential future direction of research for XRL \cite{dazeley2021levels}. \textit{Goal-driven explanation}, also referred to as \textit{eXplainable Goal-Driven AI (XGDAI)}, is an emerging area of importance in the XAI literature with recent papers surveying the concept \cite{anjomshoae2019explainable, anjomshoaeintelligible,sado2020explainable, dazeley2021levels}. This recent work shows a growing recognition that the only way people will accept an agent's behaviour is if the system provides details around the context in which its decision was based \cite{dazeley2008epistemological}. \citeauthor{langley2017explainable} \cite{langley2017explainable} describes this as \textit{explainable agency} and is considered as a first-order explanation \cite{dazeley2021levels}, where the aim is to communicate the agent's Theory of Mind \cite{leslie1987pretense}. Goal-driven explainability is primarily focused on Belief, Desire, Intention (BDI) agents \cite{rao1995bdi}, although \cite{sado2020explainable} recognises the potential for XRL in providing such explanations. In particular, \citeauthor {sado2020explainable} \cite{sado2020explainable} accepts approaches to explaining actions, discussed in section \ref{Sec:Actions}, as a post-hoc and domain independent approach to explaining behaviour. 

The difficulty is that RL does not explicitly project the effect of their actions and associate them with a goal. Therefore, when there is no model RL is essentially learning a habit, rather than a goal \cite{doll2012ubiquity}. For most applications, this distinction is trivial as there is only a single goal and the agent learns a habit for how to solve it. Beyond informing the user of what the goal is, explaining the choice of goal (when there is only one to choose from) is relatively meaningless. Therefore, for XRL to provide meaningful goal explanation it should have multiple goals that it could be pursuing at any given time. This utilisation of multiple goals, while not part of the standard RL framework, is a well-established approach with extensions to RL such as hierarchical \cite{Al2015hierarchicalRL, barto2003recent,pezzulo2018hierarchical, botvinick2009hierarchically, botvinick2012hierarchical}, multi-goal \cite{Plappert2018MultiGoalRL, vattam2013breadth, Teh2017MultitaskRL, Kaelbling93MultitaskLearning}, and multi-objective \cite{roijers2013survey, vamplew2011empirical, vamplew2018human}. This paper argues that more  meaningful goal-based explanations can be provided if RL utilise these methods more readily. 

As seen in section \ref{Sec:Actions}, the first attempts to utilise MORL to provide contrastive explanations \cite{sukkerd2018toward, sukkerd2020tradeofffocused, erwig2018explaining, erwig2020explanations} have been published. The aim for a goal-based explanation though would be to extend this initial work and answer questions about the XRL-goal being selected and how that goal affected the action selection. For instance, \citeauthor{karimpanal2017identification} \cite{karimpanal2017identification} identifies `interesting states' and learns how to find them using off-policy learning while focusing on its primary objective. Attaching a goal-based explanation to this would allow explanation about how actions could also lead to/or avoid alternative objectives. A second example would be when an agent is performing a primary task, but has an alternative objective to avoid dangerous situations \cite{vamplew2021potential}, then an explanation can identify contrastive explanations for an action on the basis of the primary or secondary objective, e.g. "While X was the fastest action, I chose Y because it was safer". 

Multi-goal \cite{Plappert2018MultiGoalRL, vattam2013breadth, Teh2017MultitaskRL, Kaelbling93MultitaskLearning} and Hierarchical \cite{Al2015hierarchicalRL, barto2003recent,pezzulo2018hierarchical, botvinick2009hierarchically, botvinick2012hierarchical} RL provide mechanisms for identifying alternative or sub-goals to problems and switching or progressing through these during a problem-solving process. At this stage there does not appear to be any attempts to provide explanations based on the currently selected goal as a means of providing better contextual information to a user. However, this paper has suggested the provision of such explanations would be a valuable area of pursuit. For instance, \citeauthor{beyret2019dot}'s \cite{beyret2019dot} approach could be extended to provide an explanation for the currently active goal through a tree traversal of potential goals using way-points during the inference process.


\subsection{Explanation of Disposition}
\label{Sec:Dispositions}

Agents that are changing their goals and/or objectives do not generally do so randomly. Some agents may do so because they have learnt that a sequence of sub-goals are required to achieve its primary goal \cite{barto2003recent,botvinick2012hierarchical, gebhardt2020hierarchical, zhou2020temporal}. Others may have multiple conflicting objectives \cite{roijers2013survey}, such as achieving a task while maintaining a safe working environment \cite{vamplew2018human, vamplew2020potential}. This process of changing goals or objectives is a result of variations made in an agent's internal disposition \cite{dazeley2021levels}. It is important that an agent is, therefore, able to include in its explanation how its current internal disposition has influenced the current choice of goal or objective. 

This change can be caused by: an observation that the prior goal was no longer appropriate for achieving its primary goal; an observed change in the environment, possibly by an external actor; or, a change in an internal simulation of an emotion, belief or desire. In Cognitive Science the Theory of Motivated control investigates how behaviour is coordinated to achieve meaningful outcomes \cite{shapiro1996controlling}. In particular, \citeauthor{pezzulo2018hierarchical} \cite{pezzulo2018hierarchical} discusses the multidimensional and hierarchical nature of goals when decision making. Essentially, people weigh-up conflicting objectives through a hierarchy of goals \cite{pezzulo2018hierarchical}. Through careful introspection it is possible for an RL agent to identify these changes in its internal disposition and provide an explanation for these changes. Such an explanation would represent a first-order explanation \cite{dazeley2021levels} and provide a valuable insight into an agent's reasoning for a human observer.

Currently there have not been any examples of explaining such dispositional RL systems, but there are numerous examples of agent-based systems, including RL, that adapt their goal autonomously during their operation. Intrinsically motivated RL has been researched for two decades, where agents construct a hierarchy of reusable skills dynamically \cite{barto2004intrinsically, chentanez2005intrinsically, chentanez2005intrinsically}. These agents change their operating goal due to internal changes such as motivations \cite{kulkarni2016hierarchical}. While methods such as \citeauthor{beyret2019dot} \cite{beyret2019dot} explain an action relative to a goal, they could be extended to explain the motivation behind its choice of goal and skill.

Disposition and motivation is not just hierarchical, but also multidimensional \cite{pezzulo2018hierarchical}. For instance, \citeauthor{vamplew2017steering} \cite{vamplew2017steering} and \citeauthor{vamplew2015reinforcement} \cite{vamplew2015reinforcement} used an algorithm referred to as Q-steering to provide the agent the ability to switch between objectives autonomously. When objectives are in conflict the agent can have an internal desire to focus on one over another and while it pursues that objective the desire to switch to an alternative objective often increases until that change is made. This approach has potential in several domains where autonomous balancing of objectives is required. An explanation for identifying the reason behind switching between policies would provide a user valuable information.

The recently emerging research in Emotion-aware Explainable AI (EXAI) methods illustrate an interest in providing explanations for agent's internal dispositions \cite{kaptein2017role}. This work focuses on self-explaining emotions and can identify important beliefs and desires. While this work is based in on a BDI framework, \citeauthor{dazeley2021levels} \cite{dazeley2021levels} argues that this can be extended to XRL. One example of this approach in RL is \citeauthor{barros2020Moody} \cite{barros2020Moody} which uses \citeauthor{cruz2020XRobot}'s \cite{cruz2020XRobot} introspection-based approach to identify an explanation, which is used to provide a self-explanation so that it can self-determine its intrinsic `mood' concerning its performance in competitive games. This approach uses an explanation informs the agent's behaviour directly. However, \citeauthor{barros2020Moody}'s \cite{barros2020Moody} approach does not currently provide an explanation for how this dispositional change has affected its current goal. Currently providing such an explanation is not evident in the XRL literature and represents an opportunity for future research. 

\subsection{Explanation of Events}
\label{Sec:Events}

In many real world applications, an RL agent will be required to deal with stochastic and dynamic environments \cite{wiering2001reinforcement}. In such environments unplanned events will occur potentially creating unexpected outcomes. An explainable agent, in such an environment, will be expected to explain how that event caused an outcome, or provide a full causal path detailing how the event caused any changes in the agent's disposition, goal or action selection. For an agent to provide such an explanation it must be able to predict the future states that would arise independent of the presence and actions of other actors within the environment. The agent's response in terms of disposition, goals and actions of the expected state and the actual state can then be compared to provide such an explanation. An extension of this model would also be able to explain what the event was that changed the environment from that which was expected. Therefore, it must be able to model the nature of stochastic events or model external actor's behaviour to understand how they may affect the environment. Therefore, this type of explanation requires the agent to perform a second-order, or social, level of explanation \cite{dazeley2021levels}. 

There are a range of value-based approaches to optimising an agent's behaviour in such environments. For instance, Robust RL \cite{morimoto2005robust}, and specially designed training mechanisms \cite{pieters2016q} can provide value-based solutions for learning and adapting in stochastic and dynamic environments. However, these approaches rarely predict the future state or model changes in the environment explicitly. Therefore, providing an explanation facility with such approaches is unlikely to provide suitable results. The nature of requiring an explicit prediction for such an explanation excludes the direct application of value-based RL methods without some form of separate predictive model being developed. For instance one approach used for this is the utilisation of generative adversarial networks (GANs) \cite{aissa2020survey, gui2020review, goodfellow2014generative} and even recurrent generative adversarial networks (RGANs) \cite{mardani2017recurrent, salehinejad2017recent}. In RL these methods are more frequently being referred to as Predictive State Encoders \cite{venkatraman2017predictive, gregor2018temporal} and are used to generate future states, also called belief states, and to predict dynamic actor's behaviour \cite{gupta2018social, kuefler2017imitating}. Similarly, in model based methods there has been significant work in developing multiple models of a domain, where prediction errors are used to select the controller or policy \cite{doya2002multiple, clavera2018model, vuong2019uncertainty}. 

There is evidence of this being a valuable form of explanation by work in BDI agents \cite{neerincx2018using, rao1995bdi}. As the name suggests, BDI agents use knowledge engineering principles to explicitly model beliefs, desires and intentions for an agent. Using knowledge-based graph traversals the beliefs about events and external actors can be a component of an integrated broad explanation of the system's behaviour. Similarly, outside of BDI research, there have been planning methods developed for providing such an explanation. Based on knowledge engineering principles, these approaches utilise abductive reasoning to generate explanations \cite{molineaux2011just, friedman2011constructing}. \citeauthor{molineaux2011learning} \cite{molineaux2011learning} presents a particularly interesting method that learns an event-model to explain anomalies through generative abductive reasoning over historical observation in partially observable dynamic environments. Currently explaining events in stochastic and dynamic environments has not been done in the RL space. As XRL research moves away from debugging style explanation towards non-expert focused explanations, providing event-based explanations is an important future research direction.


\subsection{Explanation of Expectations}
\label{Sec:Expectations}

Traditional RL has a single goal that is generally defined by the reward engineer implementing the solution. This goal is an articulation of the expectation being placed on the agent. Due to the `hard coded' nature of this expectation there is little need to explain how this expectation has caused its behaviour. However, this approach only allows for the development of very narrowly defined agents and is not applicable as agents become increasingly societal integrated with society. Such a system must adapt to their dynamic surroundings; changing their disposition and goal based on the cultural expectations of the external society with which they are integrated. Any such system must also be able to explain what expectations it is using to modify its behaviour. \citeauthor{dazeley2021levels} \cite{dazeley2021levels} extensively discussed these N\textsuperscript{th}-order explanations and the need for an autonomous agent operating in a human-AI integrated environment to model the cultural expectations that other actors may have on how the agent should behave. 

Expectations may be easily codifiable rules such as government enforced laws, military rules of engagement, ethical guidelines or business rules or they can be more abstract, learnt, or niche rules such as staying out of the doctor's way when they are rushing through an emergency ward. To meet these expectations an agent is required to change their behaviour away from their primary objective, whatever that might be. These changes in behaviour represent an area that must be explained as it may not be obvious to observers why an agent behaved in the way that it did. In particular, it should be able to articulate what expectation the agent is pursuing at any given time, why it selected that expectation, and how that changed its behaviour. 

Only agents that actively maintain a model of the expectations being placed upon it would require such explanations and currently this can only be done in RL through the incorporation of secondary systems. For instance, behaviour modelling has been studied in several fields such as BDI based Normative Agents \cite{adam2016bdi, santos2017detection, hollander2011current, beheshti2014normative}; Game Theory \cite{myerson2013game, camerer2011behavioral, suleiman2012tools, silver2016alphago, silver2016mastering}; Emotion-Driven or Emotion Augmentation learning \cite{marinier2008emotion, elliott1998model, marinier2009computational, hoey2016affect, gadanho2001robot, hao2019emotion, yu2019emotion}; and, most directly by Social Action research, which models the external demands placed on an agent that affect its goals or actions \cite{castelfranchi1998modelling, conte2016cognitive, poggi2010cognitive}. Direct use of expectation in RL is evident where some systems are designed to incorporate social and cultural awareness in to their action selection mechanism, such as pedestrian and crowd avoidance systems \cite{charalampous2017recent, chen2017socially, triebel2016spencer, kim2016socially, vasquez2014inverse, ritschel2018socially}. 

Like explanations of events, there is currently no known examples of XRL research into providing explanations of such systems. One particularly interesting recent study by \citeauthor{kampik2019explaining} \cite{kampik2019explaining} uses the idea of Explicability \cite{kulkarni2019explicable}, where an agent can perform actions and make decisions based on human expectations. \citeauthor{kampik2019explaining} \cite{kampik2019explaining} developed an approach and taxonomy for sympathetic actions that incorporate a utility for socially beneficial behaviour at the detriment of the agent's own personal gain. This system then provided explanations for the agent's behaviour resulting from these expectations. Furthermore, \citeauthor{kampik2019explaining} \cite{kampik2019explaining} recognises the relevance and applicability of this approach to RL based systems. Identifying papers in this space is, however, difficult as there is no defined research domain for this research and papers are often published under more generic fields such as understandability \cite{hellstrom2018understandable}, transparency \cite{wortham2017robot}, predictability \cite{dragan2013legibility}.


\section{Using the Causal Explainable Reinforcement Learning Framework}
\label{Sec:Using}

The work discussed previously focused on how each of the individual components of the Causal XRL Framework (CXF) has, or could be, implemented. This section will briefly look at how the CXF can be implemented and used. To some extent this can simply involve implementing all of the approaches in a single explanation facility for an agent. For instance, a system could initially use a technique such as \citeauthor{greydanus2017visualizing} \cite{greydanus2017visualizing} to identify the active features of a state. These key features could then also be used to learn causal links between actions and outcomes creating a model similar to those developed by  \citeauthor{madumal2019explainable} \cite{madumal2019explainable}, \citeauthor{khan2009minimal} \cite{khan2009minimal} or \citeauthor{cruz2019XRL} \cite{cruz2019XRL}. The combination of these approaches could provide answers to many of the reactive explanations required of the Simple-CXF. Extending these approaches to incorporate multiple objectives \cite{sukkerd2018toward, sukkerd2020tradeofffocused}, or reward decomposition \cite{erwig2018explaining, erwig2020explanations} would allow expressive contrastive and counterfactual explanations that would also facilitate the explanation of the goals and dispositions behind those choices. Second and N\textsuperscript{th}-order explanations of events and expectations would require an agent to construct models of other actors in dynamic environment using approaches such as Predictive State Encoders \cite{venkatraman2017predictive, gregor2018temporal}, Emotion Augmentation \cite{marinier2008emotion, elliott1998model, marinier2009computational, hoey2016affect, gadanho2001robot, hao2019emotion, yu2019emotion}, or Social Action \cite{castelfranchi1998modelling, conte2016cognitive, poggi2010cognitive}. Methods would then need to be developed to explain how these models affected the agent's expectations, disposition or its interpretation of an event. Such an approach combining all these elements would accomplish the idea of explaining the full details of the decision.

One important example illustrating the potential of this combined approach was a study of non-experts carried out by \citeauthor{anderson2019mental} \cite{anderson2019mental}. This study of 124 participants found that there was a significant improvement in the explainee's mental model of the agents behaviour when both XRL-Perception and XRL-Behavioural explanations were provided, when compared to only providing one or no explanation. This suggests that the combined approach was of value in improving peoples understanding. However, \citeauthor{anderson2019mental} \cite{anderson2019mental} also found that the combined explanation created disproportionately high cognitive loads for the explainee. This suggests that providing explanations across all categories would be unwieldy and difficult for most people to understand --- simply because there is too much, potentially conflicting, information. This results aligns with \citeauthor{lombrozo2007simplicity}'s \cite{lombrozo2007simplicity} suggestion that an explanation needs as few causes (simple) that cover as many events (general) and maintain consistency with peoples' prior knowledge (coherent) \cite{thagard1989explanatory}. Therefore, simply merging explanation facilities brings us no closer to presenting explanations that improve understanding, and hence trust and social acceptance. 

\begin{figure*}
  \centering
  \includegraphics[trim={0.4cm 4.1cm 0.6cm 4.0cm},clip,width=30pc]{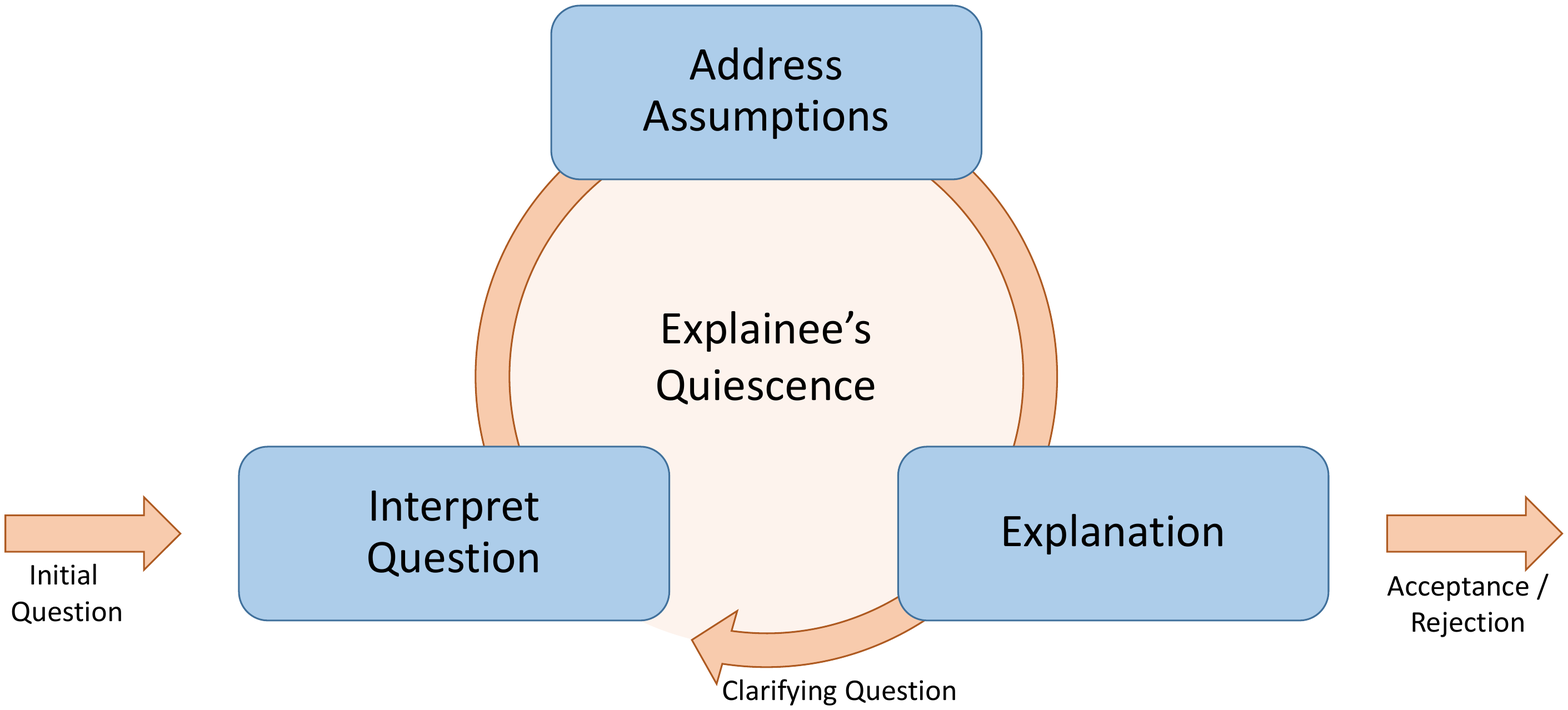} 
  \caption{A conversational model for explanation, where an agent iterates through three stages until the explainee is satisfied. The agent starts at the highest level of explanation and progresses down the pyramid, Figure \ref{fig:Levels} to more specific explanations. This conversational model is from \citeauthor{dazeley2021levels} \cite{dazeley2021levels}.}
  \label{fig:Quiescence}
\end{figure*}

\citeauthor{dazeley2021levels} presents a model for conversational interaction for explaining an AI agent's behaviour, reproduced in Figure \ref{fig:Quiescence}. The proposed model suggests that the agent presents the explanations incrementally over a sequence of interaction cycles. It is suggested that such a model would start at the highest level of intentionality in its explanation (N\textsuperscript{th}-order) and progress down the pyramid, Figure \ref{fig:Levels}, until the explainee reaches a point of \textit{Quiescence} (state of being quiet) representing a measure of stability in the user's understanding and acceptance and no longer requires deeper explanations. This model of conversational explanation aligns with the CXF proposed in this paper. Due to each of the CXF categories being aligned to the levels of explanation \cite{dazeley2021levels}, this paper proposes that an implementation of the CXF can use this model to break the range of explanation types down and only present those that are required for the user at that time. 

In this model the user would initially pose a query concerning an agent's decision, either explicitly or implicitly, which is first interpreted. The second stage attempts to identify and clarify any assumptions. This stage allows the agent to skip higher levels of explanation and go straight to the lower level explanations to address any assumptions if required. For example, if the user asks: `why didn't you catch the ball?' There is an assumption that the agent was aware that there was a ball, or that it did not succeed in catching the ball. Therefore, in resolving such assumption the agent should first determine if it was aware of a ball, and secondly, whether the outcome was in fact that no ball was caught. In the event that the assumptions are incorrect, e.g. there was no ball in its perception, then the explanation provided in the last stage skips the higher levels and provides the relevant lower level explanation. If there are no assumptions or if they are correct then the agent provides the highest level explanation as this is the most general. In the final stage the agent, using ontological models or visualisations etc, to provide a causal explanation at the determined level. 

This approach ensures that the explanation is coherent, focused on the explainee's context, while otherwise being as general as possible. In the event that the explanation does not satisfy the user they will either ask a follow up question, or through body language, indicate they are not satisfied. In such situations, the agent simply progresses to the next lower level explanation. The process ends once: the user expresses satisfaction; they change their questions to a new topic; or, all available explanations have been provided. This interactive approach to communicating explanations to a user represents a process where the agent aims to facilitate the development of a shared mental model with a human. This shared mental model is key in many situations, particularly in team-based and socially integrated domains. Development of these shared mental models has been previously explored by \citeauthor{tabrez2019improving} \cite{tabrez2019improving} where an agent uses a process referred to as Reward Augmentation and Repair through Explanation (RARE), based on inverse RL, to infer the most likely `reward' function used by a human collaborator and explain how that differs from the optimal function. In other words this project, is similar to this paper's approach in that it is providing an explanation in the context of the explainee's current understanding. 

Currently, there is no attempt to build a facility like the CXF in the XRL literature, apart from some attempts to combine perception with behaviour \cite{anderson2019mental, madumal2019explainable} and suggested extensions to combine actions with goals \cite{cruz2020XRobot, madumal2019explainable, sukkerd2018toward, sukkerd2020tradeofffocused}. The approach has been extended more thoroughly outside of XRL, using generic explanation facilities. These systems observed interactions by the agent and use the learnt model to provide explanations, such as Local Interpretable Model-Agnostic Explanations (LIME) \cite{ribeiro2016should} and Black Box Explanations through Transparent Approximations (BETA) \cite{lakkaraju2017interpretable}. Both of these approaches provide explanations across a subset of the components in the CXF.

One particularly notable example is \citeauthor{neerincx2018using} \cite{neerincx2018using}, which extended LIME and separated perceptual explanations from the cognitive processing to provide holistic explanations. The cognitive processing component incorporated goal and dispositional explanation based on emotion-based explanations. Finally, the approach incorporated ontological and interaction design patterns to communicate explanations. This approach represents the most advanced implementation utilising \cite{dazeley2021levels} levels of explanation based on intentionality and could be interpreted using multiple sections of the CXF. 

\section{Conclusion}
\label{Sec:Conclusion}

Reinforcement Learning (RL) is widely acknowledged as a one of three subfields of Machine Learning, where an agent learns through interaction with the environment using trial-and-error. However, research in the eXplainable RL (XRL) is often published under the area of Interpretable Machine Learning (IML), along with supervised learning approaches to explanation. This categorisation, however, misrepresents the possibilities that XRL presents. This paper's aim was to articulate how XRL is distinct when compared to IML, and that it offers the potential to go well beyond simply interpreting decisions. More importantly, that XRL could be the foundation to the development of truly \textit{Broad-XAI} \cite{dazeley2021levels} systems that are capable of providing trusted and socially acceptable AI systems to the wider public. In order to illustrate this the paper's second contribution was to provide a conceptual framework, referred to as the Causal XRL Framework (CXF), that highlights the range of explanations that can be provided. This framework was used to review the current extent of research that has been carried out and to identify opportunities for future research.

The Causal XRL Framework (CXF), presented in Figure \ref{fig:Framework} is based on the Casual Explanation Network (CEN) suggested by \citeauthor{bohm2015people} \cite{bohm2015people}. The CEN presents a cognitive science view of how people explain behaviour and extends prior work in attribution theory \cite{schank2013scripts}. Like the CEN, the CXF identifies seven components to causal thinking about an actor's behaviour. This directed graph of causal relationships includes a single sink node representing the outcome. This outcome is caused by either an intentional action by an agent, or from an unintended or uncontrolled sequence of events. These events could be a result of stochastic actions or external actors. The agent's actions is caused by a goal which in turn may be altered by its internal and temporary disposition, such as a change in parameter, simulated emotion or safety threshold being passed. Finally, the disposition can be affected by external cultural expectations placed upon the agent or by its perception of the world. A simplified framework, referred to as the Simplified-CXF, containing only perception and action causes for an outcome, was also provided, which represents the most current research in XRL.

In surveying the current state of the art research into XRL this paper discussed how most of the XRL-Perception was derived directly from IML research. This conection with IML is due to RL's utilisation of standard supervised learning approaches for function approximation and state feature extraction. However, there were also several examples moving beyond straight IML; providing both model and value-based extensions that were specific to XRL. These methods use introspection of the RL framework to identify causal relationships between the perceived features and either the action selected or the outcome that resulted. Of particular interest was the emergence of methods used for generating counterfactual and contrastive explanations based on these causal links between state-features and outcomes.

XRL-behaviour, as a subbranch of XRL where the agent's choice of action and the effect it has on the outcome, was explored in detail. In particular, the temporal nature of RL allows causal explanation over a sequence of actions. In surveying the literature, a number of approaches were identified. One branch of extensive work research in this space is model-based RL methods. These methods extended historical domains like Bayesian Networks and Decision Networks in providing dynamic explanations. A second branch that has seen recent results is using introspection in value-based RL, where agents either learn a transition model during interaction with the environment or convert internal values into meaningful explanations.

Finally, in discussing the full framework, this paper identified several opportunities for further research in XRL, such as using hierarchical, multi-goal, multi-objective and intrinsically motivated RL techniques in Goal-driven explanation and emotion-aware explainable AI (EXAI). This paper also discussed hypothetical approaches to the development of event-based and expectation-based explanations, such as utilising predictive state encoders and explicability in RL. Currently, these areas have been studied by other fields of explanation, such as BDI agents and Social Action, but represent the fringe of XRL research. This paper suggests these are exciting areas of future study that this field should pursue so that RL can be more widely used in real-world human-agent mixed application domains. 


\bibliographystyle{cas-model2-names}

\bibliography{cas-refs}

\balance





\end{document}